\documentclass[10pt,twocolumn,letterpaper]{article}

 \usepackage[pagenumbers]{cvpr} % To force page numbers, e.g. for an arXiv version

\definecolor{cvprblue}{rgb}{0.21,0.49,0.74}
\usepackage[pagebackref,breaklinks,colorlinks,allcolors=cvprblue]{hyperref}

\def\paperID{*****} % *** Enter the Paper ID here
\def\confName{CVPR}
\def\confYear{2025}

\title{ContextFlow: Training-Free Video Object Editing via Adaptive Context Enrichment}

\usepackage{multirow}

\author{
Yiyang Chen\textsuperscript{1}, Xuanhua He\textsuperscript{2,$\dagger$}, Xiujun Ma\textsuperscript{1,$\dagger$}, Yue Ma\textsuperscript{2,$\dagger$}, \\
\textsuperscript{1} State Key Laboratory of General Artifical Intelligence, Peking University, Beijing, China \\ 
\textsuperscript{2} The Hong Kong University of Science and Technology \\
Project Page: {\color{magenta}https://yychen233.github.io/ContextFlow-page}
}

\begin{document}
\twocolumn[{%
\renewcommand\twocolumn[1][]{#1}%
\maketitle
\vspace{-0.7cm}
\includegraphics[width=\linewidth]{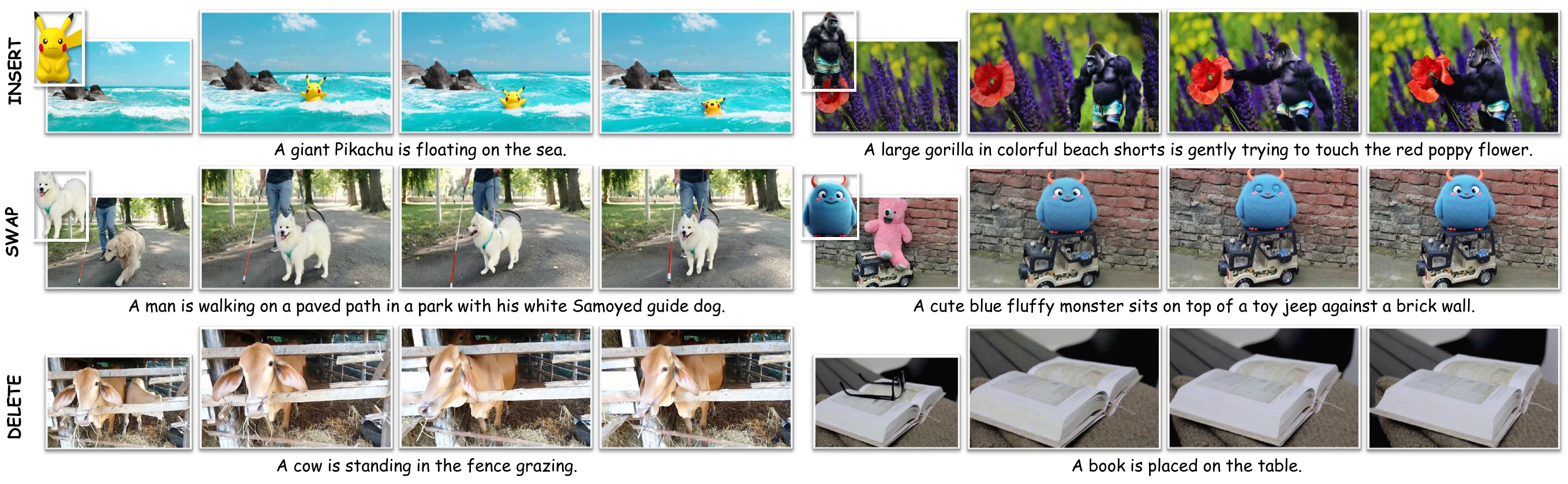}
}
\vspace{-0.2cm}
\captionof{figure}{\textbf{Showcase of ContextFlow}. Our ContextFlow achieves versatile and high-fidelity video object editing without any training. Our method demonstrates superior ability in a range of object-related challenging tasks, including object insertion (1st row), swapping (2nd row), and deletion (3rd row). The core design of our approach is Adaptive Context Enrichment, which allows for seamless integration of new elements with realistic interactions and meticulous preservation of the original scenes. 
\vspace{2em}
\label{fig:shot}
}
]

\def\thefootnote{$\dagger$}\footnotetext{Corresponding Authors.
}
\begin{abstract}
Training-free video object editing aims to achieve precise object-level manipulation, including object insertion, swapping, and deletion. However, it faces significant challenges in maintaining fidelity and temporal consistency. Existing methods, often designed for U-Net architectures, suffer from two primary limitations: inaccurate inversion due to first-order solvers, and contextual conflicts caused by crude ``hard'' feature replacement. These issues are more challenging in Diffusion Transformers (DiTs), where the unsuitability of prior layer-selection heuristics makes effective guidance challenging. To address these limitations, we introduce ContextFlow, a novel training-free framework for DiT-based video object editing. In detail, we first employ a high-order Rectified Flow solver to establish a robust editing foundation. The core of our framework is Adaptive Context Enrichment (for specifying \textit{what} to edit), a mechanism that addresses contextual conflicts. Instead of replacing features, it enriches the self-attention context by concatenating Key-Value pairs from parallel reconstruction and editing paths, empowering the model to dynamically fuse information.
Additionally, to determine where to apply this enrichment (for specifying \textit{where} to edit), we propose a systematic, data-driven analysis to identify task-specific vital layers. Based on a novel Guidance Responsiveness Metric, our method pinpoints the most influential DiT blocks for different tasks (\textit{e.g.}, insertion, swapping), enabling targeted and highly effective guidance. Extensive experiments show that ContextFlow significantly outperforms existing training-free methods and even surpasses several state-of-the-art training-based approaches, delivering temporally coherent, high-fidelity results.
\end{abstract}    
\section{Introduction}
\label{sec:intro}
Video object editing aims to achieve a range of object-related challenging editing tasks, including object insertion, swapping, and  deletion. Unlike original video editing, video object editing requires the model to meticulously preserve the unmodified background while seamlessly integrating the edited object into the video's original motion and context. This is a task that demands high spatial and temporal consistency.

Currently, there are two primary technical paths in the research community: 
training-based and training-free methods.
(1) Training-based methods aim to build powerful and feed-forward models. Recent examples include video propagation-based models like I2V Edit~\cite{ouyang2024i2vedit}, GenProp~\cite{liu2024generativevideopropagation} and ReVideo~\cite{mou2024revideo}, as well as other architectures such as VideoAnyDoor~\cite{tu2025videoanydoor}, GetIn~\cite{zhuang2025videoaddwantvideo}, VACE~\cite{chen2024vace} and UNIC~\cite{ye2025unic}, which achieve impressive results on respective benchmarks. However, these training-based methods are limited by the prohibitive computational costs and the demand for expensive large-scale datasets.
(2) Compared with the training-based methods, 
the training-free method offers a more flexible and cost-effective alternative. Early works, like AnyV2V~\cite{ku2024anyv2v}, leverage the vast knowledge embedded in pre-trained foundation models, obviating the need for any task-specific fine-tuning. This paradigm typically relies on a foundational workflow: first, inverting the source video to a noise latent using DDIM Inversion~\cite{song2020denoising}, and then guiding the new generation via Plug-and-Play (PnP) feature injection~\cite{graikos2022diffusion}. In this process, key internal features from a reconstruction of the source video are injected into the generation process of the edited video to enforce structural consistency.

However, this established workflow faces limitations. It often struggles with fidelity, leading to artifacts, inconsistent object identity, and difficulty in preserving the original background. These issues are further amplified by the recent architectural shift from U-Nets to Diffusion Transformers (DiTs), as traditional guidance mechanisms are ill-suited for this new class of models. We provide a detailed analysis in Section~\ref{sect:motivation}.

To address these critical limitations, we propose ContextFlow, a novel training-free framework that significantly advances the editing process for DiT-based models. Instead of relying on lossy inversion and crude feature replacement, ContextFlow establishes a high-fidelity, highly reversible foundation for editing based on RF-Solver~\cite{wang2024taming}. At its core is a novel Adaptive Context Enrichment mechanism, which empowers the model to dynamically fuse information from the original video and the desired edit on a per-token basis. This dynamic approach addresses the conflict between content preservation and synthesis. To apply this guidance with efficiency and precision, we propose a systematic, data-driven Vital Layer Analysis to identify the most crucial intervention points within the DiT architecture.

Our contributions can be summarized as follows:

\begin{itemize}
\item We propose ContextFlow, a novel training-free framework that is the first to apply Rectified Flow inversion to video object editing. This establishes a high-fidelity and nearly reversible foundation that significantly reduces editing artifacts.

\item We design an adaptive context enrichment mechanism that addresses the contextual conflict of feature replacement. By concatenating Key-Value pairs, our method provides ``soft guidance'' that effectively balances the trade-off between edit fidelity and content preservation.

\item We introduce a systematic and data-driven Vital Layer Analysis to identify the most crucial blocks for context injection in DiTs. This replaces the heuristic-based layer selection of U-Net frameworks and enables targeted, efficient guidance.

\item Extensive experiments on diverse editing tasks, including object insertion, deletion, and swapping, demonstrate that ContextFlow significantly outperforms existing training-free approaches and even surpasses several state-of-the-art training-based methods.

\end{itemize}

\begin{figure}[t]
\centering
\includegraphics[width=0.48\textwidth]{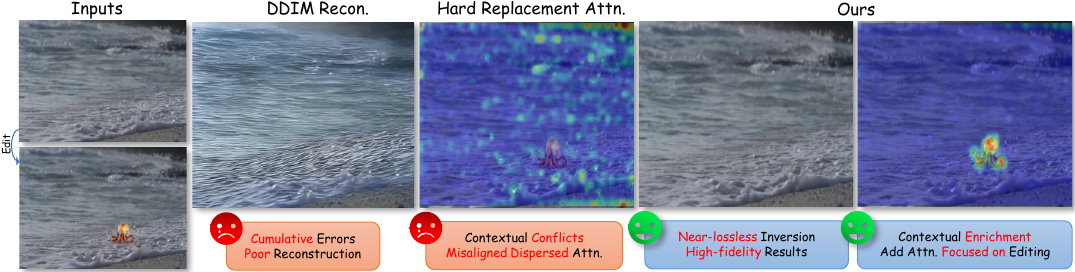} % Reduce the figure size so that it is slightly narrower than the column.
\caption{\textbf{Motivation for ContextFlow}. We highlight two core failures of prior methods: DDIM inversion causes poor reconstruction results, while ``hard replacement'' leads to misaligned attentions that only focus on the original background. ContextFlow systematically solves both.}

\label{fig:motivation}
\end{figure}
\section{Related Work}
\label{sec:ralated_work}

\subsection{Diffusion-based Video Editing}
%The landscape of video editing has been reshaped by diffusion models. The field is broadly bifurcated into tuning-based and training-free methods. Tuning-based approaches, though powerful, often incur significant computational costs by adapting pre-trained models. This includes techniques like per-video optimization \cite{wu2023tune, ouyang2024i2vedit, gao2025loraedit}, subject-driven personalization \cite{ruiz2023dreambooth, molad2023dreamix, wu2024customcrafter, wei2024dreamvideo2}, and content-motion decoupling for finer control \cite{mou2024revideo, tu2025videoanydoor}. In contrast, training-free methods pursue zero-shot editing. Early efforts focused on preserving structure by manipulating cross-frame attention \cite{geyer2023tokenflow, qi2023fatezero}. A dominant recent paradigm, however, propagates edits from a modified initial frame, leveraging strong I2V priors for impressive consistency \cite{ku2024anyv2v, bai2024uniedit}. Concurrently, other works enhance motion and appearance control by customizing the diffusion process itself \cite{jeong2024vmc, kara2024rave, wang2025consistent, burgert2025gowiththeflow}.
\begin{figure*}[ht!]
\centering
\includegraphics[width=1.0\textwidth]{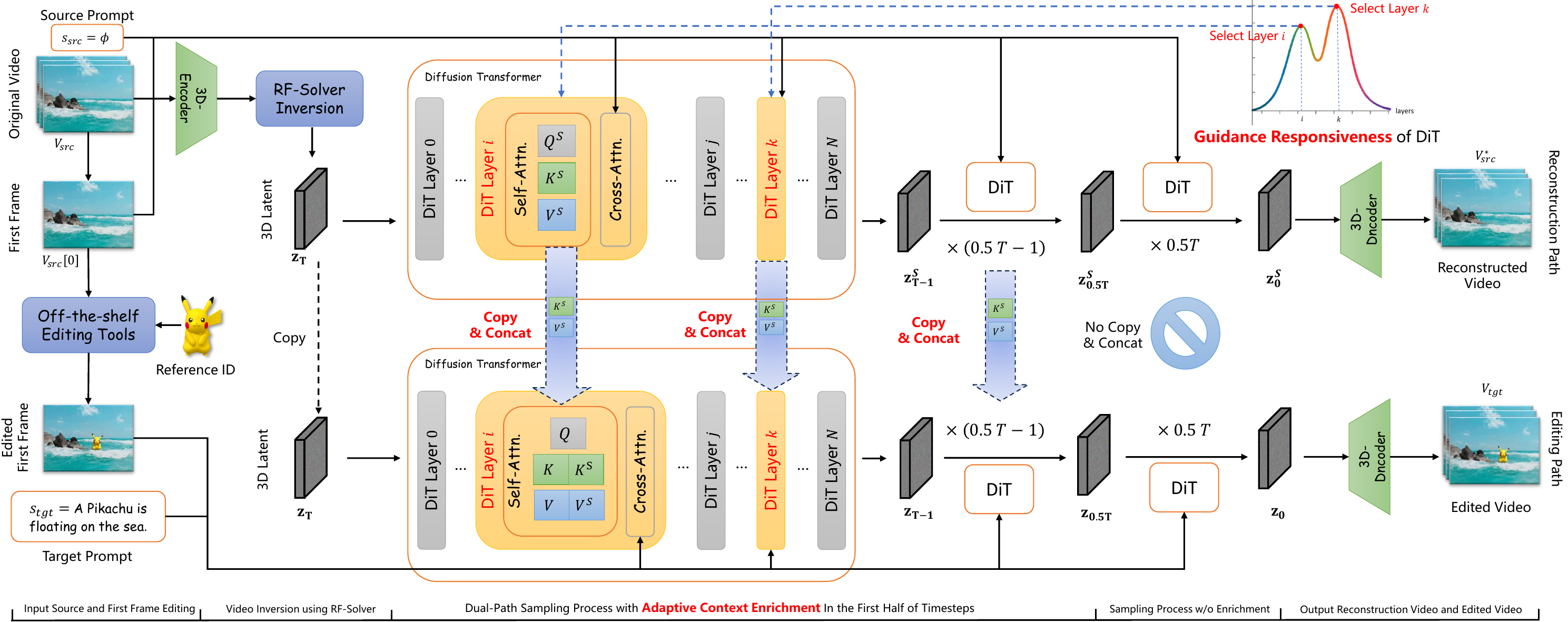} % Reduce the figure size so that it is slightly narrower than the column.
\caption{ \textbf{Overview of the ContextFlow}. Our method begins with a high-fidelity video inversion using RF-Solver to obtain a shared noise latent $\mathbf{z}_T$. A dual-path sampling process then decouples reconstruction and editing. The editing path is guided by our core mechanism, \textbf{Adaptive Context Enrichment}, where Key-Value pairs from the reconstruction path are concatenated into the self-attention blocks of the editing path. This guidance is precisely targeted to vital layers, identified via our \textbf{Guidance Responsiveness} analysis, and is only active during the first half of the denoising process to balance fidelity and consistency.}
\label{fig:overview}
\end{figure*}

The landscape of video editing has been reshaped by diffusion models. Current research in this area is largely bifurcated into tuning-based and training-free methods. Tuning-based methods adapt pre-trained models for specific videos or subjects, exemplified by per-video optimization \cite{wu2023tune} and subject-driven personalization \cite{zhu2025multibooth, zhu2024instantswap, feng2025follow, zhang2025magiccolor, yan2025eedit, ma2022visual, ruiz2023dreambooth, liu2025avatarartist, molad2023dreamix, wu2024customcrafter, wei2024dreamvideo2,he2024id,he2025fulldit2}. Other approaches focus on decoupling content and motion control during training to offer more flexible and precise editing capabilities \cite{mou2024revideo, ma2025followyourclick, ouyang2024i2vedit, ma2024followpose, ma2023magicstick, ma2022visual, ma2025controllable,che2024gamegen,zheng2024videogen,pang2024dreamdance,ma2025model}. While powerful, these methods often incur significant computational costs. Training-free methods aim for zero-shot video editing without model training. Early efforts focused on preserving video structure by manipulating cross-frame attention maps \cite{wang2024cove, geyer2023tokenflow, feng2025dit4edit, ma2025followyourmotion, ma2025followcreation, qi2023fatezero}. A significant recent trend, however, has shifted towards propagating a modified first frame by leveraging the strong generative priors of pretrained Image-to-Video (I2V) models. Seminal works in this direction, such as AnyV2V \cite{ku2024anyv2v}, have demonstrated impressive consistency and quality. This paradigm has been further generalized in frameworks like UniEdit \cite{bai2024uniedit} and GenProp \cite{liu2024generativevideopropagation}. In parallel, other works \cite{jeong2024vmc, chen2024follow, kara2024rave, ma2024followyouremoji, wang2025consistent,feng2025follow, long2025follow, burgert2025gowiththeflow} have focused on customizing the diffusion process itself to enhance motion control and appearance preservation.

\subsection{Reference-Guided Object Editing}
%To overcome the ambiguity of text-only guidance, reference-guided editing uses an image to define an object’s visual identity. A common principle is establishing feature-level correspondence to guide attention and ensure faithful appearance propagation \cite{su2025zerotohero, gu2024videoswap, ku2024anyv2v, ouyang2024i2vedit, gao2025loraedit}. A more formidable challenge is inserting a new object into a video. Novel methods tackle this by ensuring identity preservation, temporal coherence, and plausible scene interaction \cite{saini2024invi, zhuang2025videoaddwantvideo, shen2025idithoi}. Frameworks like VideoAnydoor \cite{tu2025videoanydoor} and MotionCtrl \cite{wang2024motionctrl} provide a motion prompt-based object insertion scheme to control the motion of objects in detail. Our approach aims to unify object-related video editing tasks by constructing a training-free framework based on the latest DiT model.

A critical limitation of text-only guidance is its inability to specify the unique visual identity of a particular real-world object. This has spurred the development of reference-guided editing, which uses an image to define the target’s appearance. A common principle is establishing feature-level correspondence to guide attention and ensure faithful appearance propagation. Techniques like Zero-to-Hero \cite{su2025zerotohero} establish robust feature-level connections between the reference image and video frames, guiding attention mechanisms to ensure faithful appearance transfer. This principle of using correspondence to guide propagation is shared by a variety of recent works that aim for consistent reference-based editing, including VideoSwap \cite{gu2024videoswap}, AnyV2V \cite{ku2024anyv2v}, and I2VEdit \cite{ouyang2024i2vedit}. A more formidable challenge is to seamlessly insert a new object from a reference image into an existing video. Training-free methods such as InVi \cite{saini2024invi}, GetInVideo \cite{zhuang2025videoaddwantvideo}, and iDiT-HOI \cite{shen2025idithoi} address demands that include not only identity preservation and temporal coherence but also plausible interaction with the scene. These editing-focused methods differ from reference-guided generation frameworks like VideoAnydoor \cite{tu2025videoanydoor} and MotionCtrl \cite{wang2024motionctrl}, which synthesize entirely new videos from scratch based on a reference object and motion prompts, rather than modifying pre-existing footage. Our approach aims to unify object-related video editing tasks, constructing a training-free framework based on the latest DiT model to simultaneously support multiple functions such as object insertion, object swapping, and object deletion.

\section{Method}

Given a reference video, we aim to edit objects in the video, including the object insertion, swapping, and deletion. In the following section,  we first analyze the core challenges in training free video object editing and provide our motivation in Sec.~\ref{sect:motivation}. Then the overview of the proposed ContextFlow is presented in Sec.~\ref{sec:overview}. We introduce high-fidelity inversion, adaptive context enrichment and Vital layer analysis in Sec.~\ref{sec:rfinversion}, Sec.~\ref{sec:ace} and Sec.~\ref{sec:layer_ana}.

\subsection{Core Challenges and Motivations}
\label{sect:motivation}
\subsubsection{The Challenge of Video Inversion for Editing}
For training-free editing, a critical first step is to invert a real video $V$ back to its corresponding noise latent $z_1$. This process should ideally be perfectly reversible, creating an unambiguous anchor that encodes all spatiotemporal information of the source video. This is typically modeled as solving an Ordinary Differential Equation (ODE) that describes the path from noise to data. However, in practice, this ODE is solved numerically. Standard techniques like DDIM Inversion, which rely on first-order solvers similar to the Euler method, discretize the generation path from $z_1$ to $z_v$ into N steps:
\begin{equation}
    z_{t_{i-1}} = z_{t_i} + (t_{i-1}-t_i) v_{\theta}\left(z_{t_i}, t\right)
\end{equation}
where $t_i$ goes from $1$ to $0$.
Naively reversing this process for inversion introduces significant discretization errors, which accumulate over timesteps. This results in a noisy latent that cannot faithfully reconstruct the original video, as shown in Figure~\ref{fig:motivation}, severely compromising the quality and consistency of any subsequent edits. 
\subsubsection{Contextual Conflict in Guidance} The conventional PnP mechanism, which involves a ``hard'' replacement of features, is often too crude. This rigid intervention can create a conflict between the source video’s structure and the edited object, leading to visual artifacts and inconsistent object identity. For example, queries from the editing path, seeking to form a new concept, are forced to attend to keys from the original video. This mismatch confuses the attention mechanism, leading to suppressed edits or artifacts. As visualized in Figure~\ref{fig:motivation}, hard replacement creates a semantic conflict, causing the edit-related queries to erroneously attend to keys and values from the original, irrelevant video context.  This issue is compounded by the move to homogeneous Diffusion Transformers, whose lack of distinct semantic layers---unlike hierarchical U-Nets---makes it unclear where and how to effectively inject guidance.
In response to these three core challenges, we propose ContextFlow, a novel framework designed specifically for high-fidelity, DiT-based video object editing.

\subsection{Overview of ContextFlow} 
\label{sec:overview}

Our proposed framework, ContextFlow, addresses this by designing a controlled generation process within a pre-trained I2V Diffusion Transformer, all without any weight modification. As illustrated in Figure~\ref{fig:overview}, our approach is built upon three foundational parts. First, to solve the inversion fidelity problem, we establish a near-lossless and highly reversible foundation using Rectified Flow, creating a clean canvas for editing. Next, to address the contextual conflict inherent in feature replacement, we introduce an Adaptive Context Enrichment mechanism. Instead of a crude ``hard'' replacement, this method enriches the context by concatenating the Key-Value (KV) pairs from both the source video's reconstruction and the editing path, empowering the DiT’s self-attention to dynamically balance preservation and synthesis. Finally, to answer the critical question of where to apply this guidance, our data-driven Vital Layer Analysis systematically identifies the most crucial layers for intervention. This avoids the drawbacks of naive all-layer injection and replaces unreliable heuristics, ensuring both precision and efficiency. Together, these components enable robust, high-fidelity video editing in a training-free manner.

\subsection{High-Fidelity Inversion via Rectified Flow}
\label{sec:rfinversion}
Our editing workflow begins with the source video \(V_{src}\) and a target prompt \(s_{tgt}\). Using an off-the-shelf image editor (e.g., AnyDoor~\cite{chen2024anydoor}), we first modify the initial frame \(V_{src}[0]\) to create the edited frame \(I_{edit}\). The primary challenge is then to propagate the static edit in \(I_{edit}\) throughout the video, guided by \(s_{tgt}\), while maintaining fidelity to \(V_{src}\).

As established in our preliminaries, the success of this editing hinges on the quality of the initial noise latent. Standard inversion methods like DDIM are lossy, making it difficult to disentangle editing artifacts from inversion errors. 
To eliminate this ambiguity, a higher-order numerical solver is essential. Therefore, we introduce RF-Solver~\cite{wang2024taming}, a training-free, second-order sampler that provides the highly reversible and high-fidelity mapping required for a robust generative foundation.

RF-Solver achieves its precision by utilizing a second-order Taylor expansion to more accurately estimate the ODE path during inversion.
\begin{equation}
z_{t_{i+1}} = z_{t_i} + (t_{i+1} - t_i) v_\theta(z_{t_i}, t_i) + \frac{1}{2}(t_{i+1} - t_i)^2 v^{(1)}_\theta(z_{t_i}, t_i)
\end{equation}
where \(v^{(1)}_\theta\) is the numerically estimated time derivative of the velocity field. We apply this to the VAE-encoded latents of the source video, conditioned on its original first frame \(V_{src}[0]\) and a null-text prompt \(s_{src}=\phi\), to obtain a unique noise anchor \(\mathbf{z_1}\):
\begin{equation}
 \mathbf{z_1} = \text{RF-Solver}_{\text{inversion}}(V_{src}, V_{src}[0], s_{src})
\end{equation}
The resulting anchor \(\mathbf{z_1}\) provides a faithful representation of the original video's spatiotemporal information, establishing a robust foundation for our editing mechanism

\subsection{Adaptive Context Enrichment for DiT-based Guidance}
\label{sec:ace}
With a reliable noise anchor \(\mathbf{z_1}\) established, our core challenge becomes propagating the edit from the triplet \(( \mathbf{z_1}, I_{edit}, s_{tgt} )\) while preserving the original video's structure. A naive, single-path generation, denoising from \(\mathbf{z_1}\) using only the new conditions \((I_{edit}, s_{tgt})\) is insufficient as it lacks continuous guidance from the source video, leading to content drift. While prior methods use feature injection from the original source video, their reliance on ``hard replacement'' creates a fundamental problem in DiTs.
\subsubsection{The Contextual Conflicts of Hard Replacement}
\begin{figure}[t]
\centering
\includegraphics[width=0.48\textwidth]{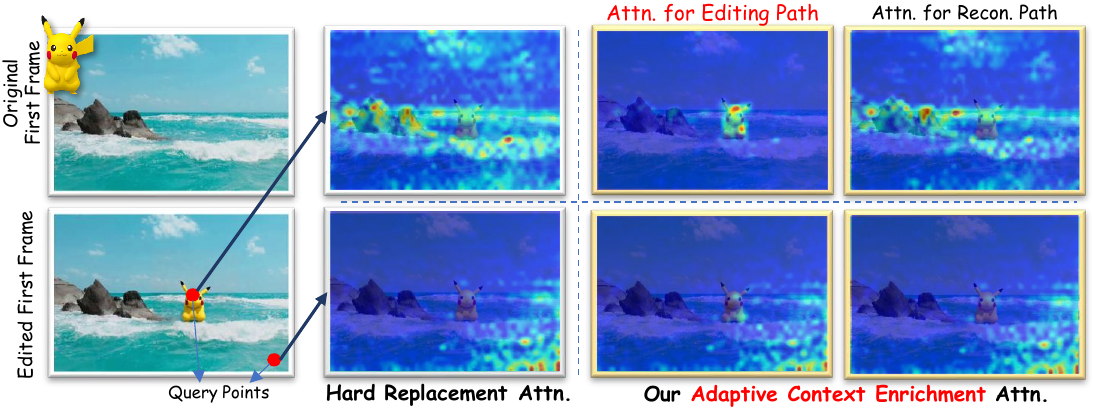} % Reduce the figure size so that it is slightly narrower than the column.
\caption{
\textbf{Resolving Contextual Conflict}. 
Hard replacement misdirects attention for edited queries, suppressing object synthesis. Our \textbf{Adaptive Context Enrichment} resolves this by offering a dual context: the Editing Path for synthesizing the new object, and the Reconstruction Path for preserving background structure. Attention in unedited regions remains correct, confirming our method is non-invasive.
%Our Adaptive Context Enrichment resolves contextual conflicts that plague conventional editing methods. When generating the inserted object (query on Pikachu, top row), Hard Replacement's attention is misaligned with the original background, suppressing the edit. Our method resolves this by allowing the query to attend to two contexts simultaneously: it focuses on the new object within the Editing Path for faithful synthesis, while concurrently referencing the original background structure (the rocks) via the Reconstruction Path. The model adaptively leverages both, enabling harmonious integration. For background queries (bottom row), attention remains faithfully on the original structure, confirming our method's non-invasive nature.
}
\label{fig:contradiction_viz}
\end{figure}

We identify the core limitation of naive feature injection as a contextual conflict. This occurs when edit-specific queries (\(Q^{edit}\)), which seek to form a new semantic concept (e.g., ``a Pikachu is floating on the sea.''), are forced to attend to a context (\(K^{res}, V^{res}\)) from the original video that only contains information about the original scene (e.g., ``the sea surface''). As illustrated in Figure~\ref{fig:contradiction_viz}, this semantic mismatch confuses the attention mechanism, leading to suppressed edits or artifacts. This necessitates a more intelligent fusion strategy.

\subsubsection{Adaptive Fusion via Context Enrichment}
Our solution is fundamentally different: adaptive context enrichment. Instead of replacing the context, we enrich it, empowering the pre-trained attention module to perform a dynamic, content-aware fusion.
To implement this, we access both editing and reconstruction contexts simultaneously via a synchronized dual-path process. Both paths originate from the same noise anchor $\mathbf{z_1}$. The first, our Reconstruction Path, is conditioned on the original video inputs $(V_{src}[0], s_{\phi})$. 
It focuses on preserving content fidelity by denoising $\mathbf{z_1}$ back to the source video, providing the essential source context (keys $K^{res}_{t,l}$ and values $V^{res}_{t,l}$). 
In parallel, the Editing Path handles the creative task. Conditioned on the edit inputs $(I_{edit}, s_{tgt})$, it synthesizes the desired changes from the same $\mathbf{z_1}$, providing the editing queries \(Q^{edit}_{t,l}\) and its own internal context \((K^{edit}_{t,l}, V^{edit}_{t,l})\).

With both sets of contexts available, we perform the enrichment within the Editing Path's self-attention. We augment the key and value by concatenating them with their counterparts from the Reconstruction Path:
\begin{align}
    K_{aug} &= \text{Concat}([K^{edit}_{t,l}, K^{res}_{t,l}]) \label{eq:k_aug} \\
    V_{aug} &= \text{Concat}([V^{edit}_{t,l}, V^{res}_{t,l}]) \label{eq:v_aug}
\end{align}

The self-attention is then computed using this expanded context:
\begin{equation}
\text{Self-Attn}_{\text{enriched}} = \text{softmax} \left(\frac{Q^{edit}_{t,l}(K_{aug})^T}{\sqrt{d}} \right)V_{aug} \label{eq:attn_enriched}
\end{equation}

This design leverages the inherent optimization behavior of self-attention. The enriched key-value space allows each query to attend to its most relevant information—be it from the source context for background preservation, or the edit context for new content synthesis. This transforms guidance from a rigid command into a dynamically weighted fusion process, enabling a robust and high-fidelity fusion.

\subsection{Targeted Fusion via Vital Layer Analysis}
\label{sec:layer_ana}
\begin{figure}[t]
\centering
\includegraphics[width=0.5\textwidth]{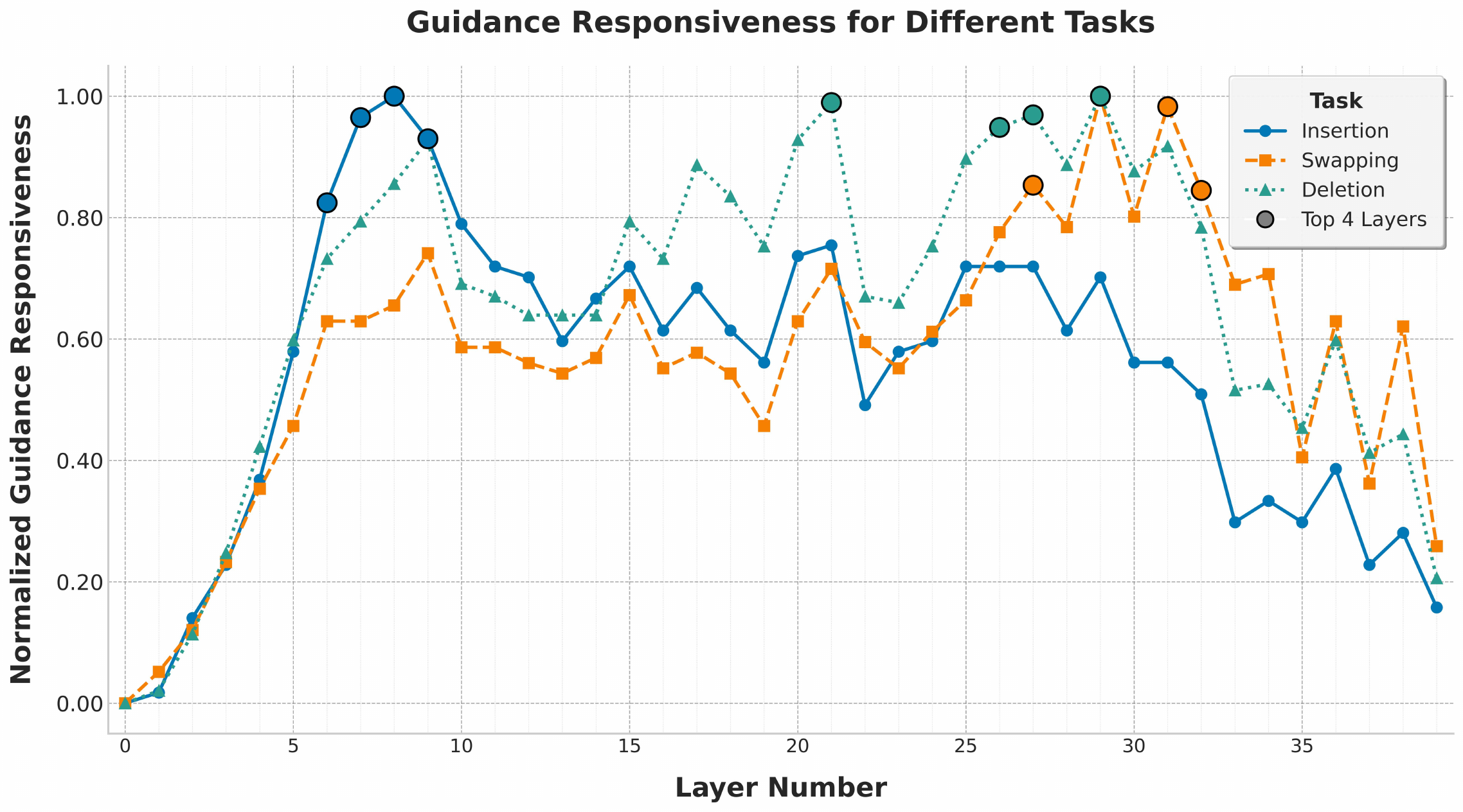} 
\caption{\textbf{Task-Dependent Guidance Responsiveness} (min-max normalized data in the figure). A higher Guidance Responsiveness indicates greater influence. There are three primary zones across all layers, which distribute in the shallow area (layers ~1-10), mid-layer area (layers ~15-21) and deep area (layers ~26-32) respectively. Moreover, the numerical ranking of Guidance Responsiveness for these three regions varies depending on the specific task.}
\label{layer-importance}
\end{figure}

Having established \textit{how} to guide the generation, we now address the critical question of \textit{where}. Injecting guidance uniformly across all layers of a DiT is not only computationally wasteful but also conceptually flawed. Applying our semantic-level guidance uniformly across all layers risks disrupting the DiT's functional hierarchy of layers, potentially weakening the edit's results.
Therefore, a targeted intervention is required. Prior works on U-Net architectures have relied on empirical heuristics for layer selection. However, such heuristics are not reliably transferable to the different and more homogeneous structure of DiTs.

To formalize this, we propose a data-driven method to identify the most influential layers. We define a Guidance Responsiveness Metric, $GR_l$,  that quantifies a layer’s responsiveness to our Contextual Enrichment mechanism. For each layer $l$, we calculate this by performing a one-step denoising on a set of videos with the editing conditions and computing two feature maps for each layer: $x^{\text{no-CE}}_l$ (i.e., layer $l$'s self-attention output without Contextual Enrichment) and $x^{\text{CE}}_l$(i.e., layer $l$'s self-attention output with Contextual Enrichment applied only at layer $l$). The Guidance Responsiveness is measured by the dissimilarity:
\begin{equation}
    GR_l = 1 - \text{mean} \left(\text{cosine\_similarity}\left(x^{\text{no-CE}}_l,x^{\text{CE}}_l \right) \right)
\end{equation}

\begin{table*}[h!]
\centering
\resizebox{\textwidth}{!}{%
\begin{tabular}{ll|cc|cc|ccc|cc}
\toprule
\multirow{2}{*}{\textbf{Task}} & \multirow{2}{*}{\textbf{Method}} & \multicolumn{2}{c|}{\textbf{Identity}} & \multicolumn{2}{c|}{\textbf{Alignment}} & \multicolumn{3}{c|}{\textbf{Video Quality}} & \multicolumn{2}{c}{\textbf{Reconstruction Quality}} \\
\cmidrule{3-11}
& &CLIP-I ↑&DINO-I ↑&CLIP-Score ↑&Overall cons. ↑&Smoothness ↑& Dynamic ↑& Aesthetic ↑&PSNR ↑&SSIM ↑\\
\midrule
\multirow{5}{*}{\rotatebox[origin=c]{90}{\textbf{Insert}}} 
& AnyV2V & 0.5943 & 0.4053 & 0.2776 & 0.1887 & 0.9804 & 0.3077 & 0.5287 & 20.57 & 0.7055 \\
& VACE & 0.5683 & 0.3967 & 0.2569 & 0.1386 & \textbf{0.9921} & 0.3077 & 0.5724 & 18.86 & \textbf{0.9033} \\
& AnyV2V-DiT & 0.6376 & 0.4479 & 0.3060 & 0.2579 & 0.9917 & 0.3846 & 0.6145 & 26.06 & 0.8478 \\
& I2VEdit & \textbf{0.6710} & \textbf{0.4595} & \textbf{0.3124} & 0.2600 & 0.9827 & 0.3077 & 0.5846 & 26.23 & 0.8360 \\
& \textbf{Ours} & 0.6504 & 0.4566 & 0.3107 & \textbf{0.2691} & 0.9918 & \textbf{0.4231} & \textbf{0.6227} & \textbf{26.26} & 0.8575 \\
\midrule
\multirow{5}{*}{\rotatebox[origin=c]{90}{\textbf{Swap}}} 
& AnyV2V & 0.6046 & 0.5641 & 0.3210 & 0.2384 & 0.9848 & 0.0769 & 0.5739 & 21.88 & 0.6867 \\
& VACE & 0.6080 & 0.5917 & 0.3226 & 0.2412 & \textbf{0.9926} & \textbf{0.1538} & 0.6144 & \textbf{29.63} & \textbf{0.9238} \\
& AnyV2V-DiT & 0.6617 &  0.5983 & 0.3362 & 0.2597 & 0.9907 & \textbf{0.1538} & 0.6076 &  20.18 & 0.6854 \\
& I2VEdit & \textbf{0.6683} &  0.6003 & 0.3282 & 0.2595 & 0.9819 & 0.0769 & 0.5995 & 26.19 & 0.7966 \\
& \textbf{Ours} & 0.6644 & \textbf{0.6004} & \textbf{0.3391} & \textbf{0.2648} & 0.9924 & 0.0769 & \textbf{0.6176} & 22.66 & 0.7518 \\
\midrule
\multirow{5}{*}{\rotatebox[origin=c]{90}{\textbf{Delete}}} 
& AnyV2V & -- & -- & \textbf{0.2891} & 0.2170 & 0.9781 & 0.1500 & 0.5378 & 22.07 & 0.6530 \\
& VACE & -- & -- & 0.2794 & 0.1948 & 0.9889 & 0.3000 & \textbf{0.5645} & \textbf{31.57} & \textbf{0.9064} \\
& AnyV2V-DiT & -- & -- & 0.2863 & \textbf{0.2136} & 0.9886 & 0.3000 & 0.5413 & 21.14 & 0.6514 \\
& I2VEdit & -- & -- & 0.2790 & 0.2081 & 0.9816 & 0.3000 & 0.5169 & 25.44 & 0.7758 \\
& \textbf{Ours} & -- & -- & 0.2854 & 0.2111 & \textbf{0.9900} & \textbf{0.3500} & 0.5405 & 22.16 & 0.7030 \\
\bottomrule
\end{tabular}
}
\caption{\textbf{Quantitative comparison} on object insertion, swap, and deletion tasks. Our method has achieved impressive performance across numerous metrics for each task, demonstrating its comprehensiveness.}
\label{tab:main_comparison}
\end{table*}

A high $GR_l$ score signifies that the layer is highly sensitive to the guidance and thus influential in the editing process. 
Applying this analysis across a range of editing tasks reveals that layer responsiveness is not uniform, but instead exhibits a highly structured and task-dependent pattern, as shown in Figure~\ref{layer-importance}. There are three primary zones of high responsiveness across the model's depth: an early block (layers ~1-10), a middle block (layers ~15-21), and a deep block (layers ~26-32). Crucially, the dominant zone of activity varies systematically with the task. For object insertion, peak responsiveness is consistently located in the early-layer block. In contrast, object swapping elicits the strongest response in the deep-layer block. Object deletion presents a unique dual-peak pattern, showing high responsiveness in both the middle and deep blocks. These empirical evidence strongly suggests a structural pattern within the DiT's layers. The observed patterns are consistent with the broader understanding of Transformer architectures, where early layers typically handle spatial and structural information, while deeper layers manage more abstract semantic concepts~\cite{tenney2019bert, raghu2021vision}. For instance, the reliance of insertion on early layers aligns with a need to establish spatial layout, while deletion's dependence on deep layers corresponds to a high-level semantic operation.

By selecting only the top-$k$ layers with the highest importance for each task, we ensure our intervention is potent, targeted, and computationally efficient. This principled selection strategy completes our framework, delivering a robust and precise video editing solution.
\begin{figure*}[t!]
\centering
\includegraphics[width=1.0\textwidth]{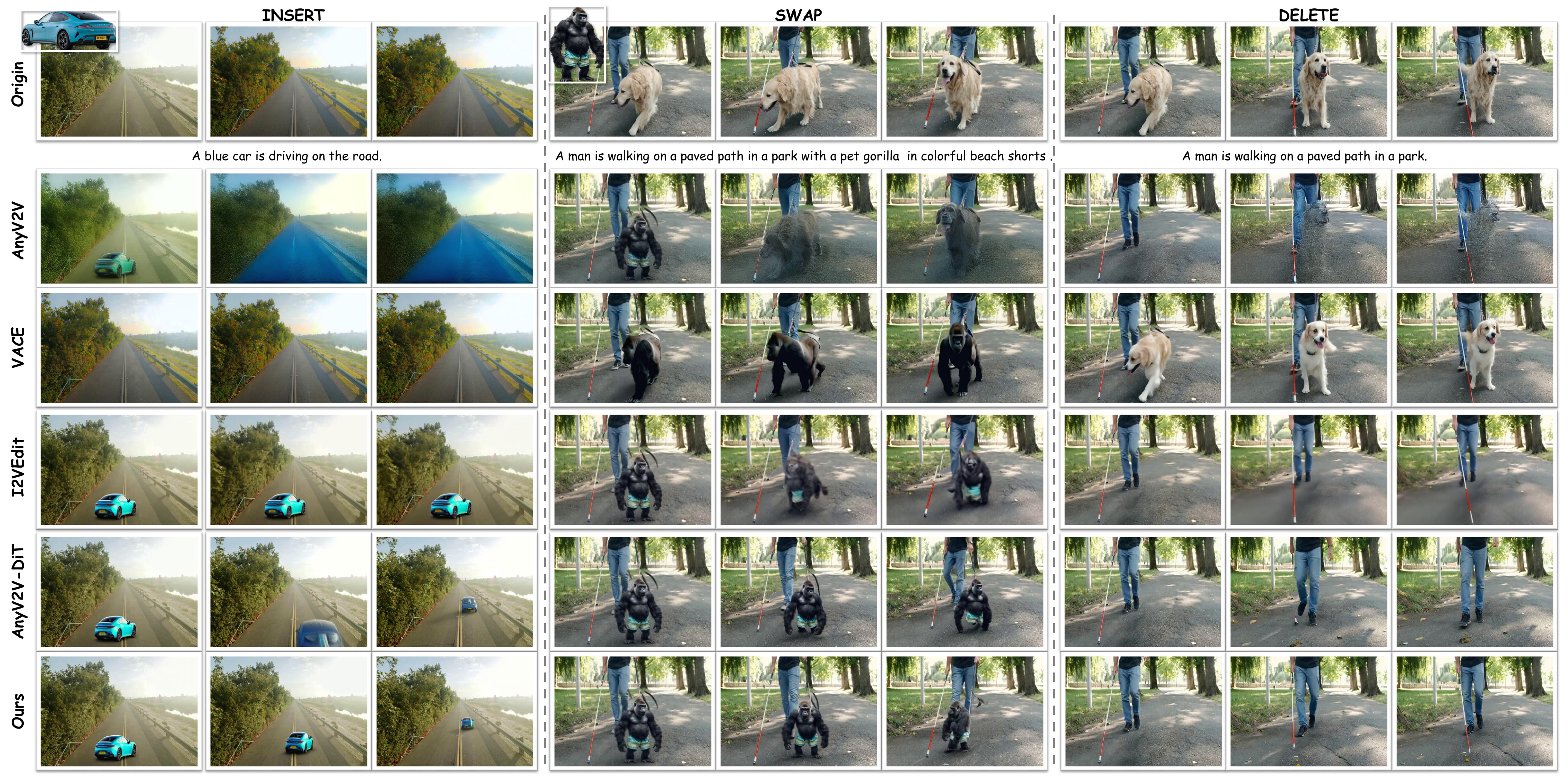} % Reduce the figure size so that it is slightly narrower than the column.
\caption{\textbf{Qualitative comparisons} of our method against open-source video editing baselines. The guiding text prompt is shown below the original videos. Our method demonstrates satisfactory results. Meanwhile, other methods either struggle to achieve the intended goals of insertion, replacement, or deletion, fail to assign proper motion patterns to the modified objects, or generate strange artifacts or low-quality videos. Zoom in for better visualization.}
\label{fig:results}
\end{figure*}

\section{Experiments}

\subsection{Experimental Setup}

\paragraph{Implementation Details.} Our framework is built upon Wan2.1-I2V-14B-480P~\cite{wan2025wan}, a publicly available image-to-video Diffusion Transformer with 40 layers. We adhere to a training-free paradigm, requiring no optimization or fine-tuning of the pre-trained diffusion model. For inversion, we utilize RF-Solver~\cite{wang2024taming} with 50 steps to map the source video into a noise latent, and the subsequent editing also uses 50 sampling steps. We set the timestep threshold $\tau$ to 0.5 (i.e., the mechanism is adopted for the first 50\% of timesteps) and a guidance scale of 3.0.

\paragraph{Evaluation Dataset and Baselines.} We evaluate our method on the Unic-Benchmark~\cite{ye2025unic}. For first-frame editing, we use AnyDoor~\cite{chen2024anydoor} for insertion, InsertAnything~\cite{song2025insert} for swapping, and MagicQuill~\cite{liu2025magicquill} for deletion. We compare against baselines including training-free AnyV2V~\cite{ku2024anyv2v}, our adapted AnyV2V-DiT, and training-based VACE~\cite{chen2024vace} and I2VEdit~\cite{ouyang2024i2vedit}. For evaluation, we measure task-specific fidelity (CLIP-score, DINO-score), background preservation (PSNR, SSIM), and overall video quality using metrics from VBench~\cite{huang2024vbench} (e.g., consistency, smoothness, dynamic and aesthetic quality).

\subsection{Comparison with baselines}

We conduct a comprehensive quantitative evaluation against state-of-the-art methods on object insertion, swapping, and deletion tasks.

The results, summarized in Table \ref{tab:main_comparison} and Figure \ref{fig:results}, demonstrate the performance of ContextFlow. For creative edits like object insertion and swapping, ContextFlow outperforms most baselines in critical Identity and Alignment metrics. Its leading Aesthetic and Dynamic scores also reflect high visual fidelity and temporal coherence. In contrast, while I2VEdit achieves higher Identity scores, it requires per-video training and produces stiff, copy-paste-like results(Figure~\ref{fig:results}). This effect is not what we desire, even though it has achieved a higher value in alignment metrics. Other methods show more significant flaws: VACE fails on out-of-distribution insertion and exhibits poor identity matching in swaps, while AnyV2V yields blurry visuals and unstable objects. 

In the object deletion task, ContextFlow achieves top scores in Smoothness and Dynamics while maintaining high reconstruction quality. In contrast, VACE incorrectly replaces objects with hallucinated content, while AnyV2V leaves shadow artifacts and shows poor reconstruction quality (Figure~\ref{fig:results}).

The ``AnyV2V-DiT'' baseline shows that directly applying traditional U-Net guidance mechanisms to the DiT architecture causes severe visual artifacts, including geometric distortions, inconsistent temporal dynamics, and spatial deformations. This result provides empirical evidence that the architectural transition to Transformers requires fundamentally new guidance mechanisms.

\subsection{Ablation Studies}
To rigorously analyze the contributions of our proposed components, we conduct a series of ablation studies on the object insertion task. Our analysis is structured to answer three fundamental questions regarding our ContextFlow framework: 1) Is our guidance mechanism effective and how should it be implemented? 2) How much guidance is optimal and where should it be injected? 3) During which phase of the denoising process should guidance be active?

\subsubsection{Core Mechanism Validation: Guidance Strategy}
We first validate the necessity and design of our adaptive Context Enrichment (CE) mechanism. We compare our method against two critical variants: one that omits Contextual Enrichment entirely, relying solely on the inverted noise and edited first frame, and another that substitutes our K/V concatenation with a conventional ``hard replacement'' strategy.

As presented in Table~\ref{tab:ablation_strategy}, ablating the CE module (w/o Contextual Enrichment) leads to a discernible decline across all metrics. While high-fidelity inversion provides a strong foundation, explicit guidance during the denoising path is crucial for context-aware editing. More revealingly, the ``K/V Replacement'' strategy significantly impairs Identity Preservation scores. We attribute this to the destructive nature of replacement, which discards valuable contextual information from the editing path's context. In contrast, our concatenation approach is additive; it enriches the context, empowering the self-attention module to dynamically balance between source and target contexts.

\begin{table}[h]
\centering
\resizebox{\columnwidth}{!}{%
\begin{tabular}{l|cc|cc|c}
\toprule
\textbf{Method} & \textbf{CLIP-I}↑ & \textbf{DINO-I}↑ & \textbf{CLIP-Score}↑ & \textbf{Overall cons.}↑ & \textbf{Aesthetic}↑ \\
\midrule
\textbf{Ours} & \textbf{0.6504} & \textbf{0.4566} & \textbf{0.3107} & \textbf{0.2691} & \textbf{0.6227} \\
\midrule
\multicolumn{6}{l}{\textit{Ablation on Guidance Strategy}} \\
\quad w/o CE. & 0.6447 & 0.4529 & 0.3086 & 0.2634 & 0.6186 \\
\quad K/V Replace & 0.6349 & 0.4508 & 0.3018 & 0.2544 & 0.6200 \\
\bottomrule
\end{tabular}
}
\caption{\textbf{Ablation on the core guidance strategy.} Our method demonstrates clear superiority over both the absence of guidance and a destructive replacement approach, highlighting the critical importance of our carefully designed strategic integration.}
\label{tab:ablation_strategy}
\end{table}

\subsubsection{Targeted Guidance Analysis: How Much and Where?}
Having established the efficacy of our mechanism, we now investigate the specifics of its application: determining the optimal \textit{quantity} and \textit{location} for K/V injection.

\paragraph{How Much Guidance?} We first analyze the impact of $\mathbf{k}$, the number of top-ranked layers selected for injection. As detailed in Table~\ref{tab:ablation_k}, the results reveal a distinct unimodal performance curve. Insufficient guidance ($k<4$) fails to provide a robust structural anchor, leading to weaker identity preservation. Conversely, excessive guidance ($k>4$) over-constrains the model, stifling its generative capacity and causing the desired edit to diminish, as evidenced by the sharp performance degradation for $k=32$ and $k=40$. An optimal balance is achieved at $k=\mathbf{4}$, which corresponds to the top 10\% of layers in the 40-layer DiT.

\begin{table}[h]
\centering
\resizebox{\columnwidth}{!}{%
\begin{tabular}{l|cc|cc|c}
\toprule
\textbf{Method} & \textbf{CLIP-I}↑ & \textbf{DINO-I}↑ & \textbf{CLIP-Score}↑ & \textbf{Overall cons.}↑ & \textbf{Aesthetic}↑ \\
\midrule
$k=0$ & 0.6447 & 0.4529 & 0.3086 & 0.2634 & 0.6186 \\
$k=1$ & 0.6467 & 0.4537 & 0.3087 & 0.2645 & 0.6196 \\
$k=2$ & 0.6452 & 0.4530 & 0.3077 & 0.2623 & 0.6188 \\
\textbf{Ours($k=4$)} & \textbf{0.6504} & \textbf{0.4566} & \textbf{0.3107} & \textbf{0.2691} & 0.6227 \\
$k=8$ & 0.6456 & 0.4541 & 0.3089 & 0.2620 & \textbf{0.6240} \\
$k=16$ & 0.6330 & 0.4435 & 0.3058 & 0.2566 & 0.6157 \\
$k=32$ & 0.6100 & 0.4303 & 0.2959 & 0.2458 & 0.5998 \\
$k=40$ & 0.5715 & 0.4067 & 0.2685 & 0.1863 & 0.5863 \\
\bottomrule
\end{tabular}
}
\caption{\textbf{Ablation on the number of injected layers.} As shown in our experiments, performance peaks at k=4, which serves as an optimal trade-off between providing sufficient guidance strength without overly constraining and ensuring proper generative freedom.}
\label{tab:ablation_k}
\end{table}

\paragraph{Where to Inject?} With the optimal quantity established as $k=4$, we now validate our principled method for selecting \textit{which} four layers to target. In Table~\ref{tab:ablation_selection}, we benchmark our Guidance Responsiveness-based selection against several alternatives. Injecting into all 40 layers proves detrimental, corroborating our earlier finding that over-constraining the model is counterproductive. Conversely, injecting into the four \textit{least}-responsive layers provides insufficient structural cues, failing to anchor the edit effectively. Finally, a U-Net-based heuristic adapted from AnyV2V (layers 3-10) performs commendably but is still surpassed by our method on key identity metrics. This demonstrates that our data-driven Guidance Responsiveness metric is not merely a theoretical construct but a practical and superior tool for identifying the most impactful layers for precise video editing.

\begin{table}[h]
\centering
\resizebox{\columnwidth}{!}
{%
\begin{tabular}{l|cc|cc|c}
\toprule
\textbf{Method} & \textbf{CLIP-I}↑ & \textbf{DINO-I}↑ & \textbf{CLIP-Score}↑ & \textbf{Overall cons.}↑ & \textbf{Aesthetic}↑ \\
\midrule
\textbf{Ours} & \textbf{0.6504} & \textbf{0.4566} & 0.3107 & \textbf{0.2691} & \textbf{0.6227} \\
\midrule
\multicolumn{6}{l}{\textit{Ablation on Layer Selection Strategy}} \\
\quad Injection on All layers & 0.5715 & 0.4067 & 0.2685 & 0.1863 & 0.5863 \\
\quad Inj. on 4 least-responsive & 0.6138 & 0.4367 & 0.2970 & 0.2365 & 0.6059 \\
\quad Inj. selection follow AnyV2V & 0.6458 & 0.4553 & \textbf{0.3109} & 0.2634 & 0.6210 \\
\bottomrule
\end{tabular}
}
\caption{\textbf{Ablation on the layer selection strategy.} Our principled selection consistently outperforms alternative approaches, clearly underscoring the notable effectiveness of our proposed Guidance Responsiveness metric and its strong correlation with performance.}
\label{tab:ablation_selection}
\end{table}

\subsubsection{Guidance Window Analysis: When to Inject?}
Finally, we analyze the injection timestep threshold $\tau$, which governs the temporal duration of the Contextual Enrichment mechanism. This parameter critically mediates the trade-off between structural preservation (favoring higher $\tau$) and edit flexibility (favoring lower $\tau$). As evaluated in Table~\ref{tab:ablation_tau}, a low $\tau=0.2$ offers insufficient guidance, while a high $\tau=1.0$, which applies guidance throughout the entire process, slightly compromises aesthetic quality by restricting the model during the final, high-fidelity refinement stages. We identify $\tau=0.5$ as the optimal equilibrium.

\begin{table}[h]
\centering
\resizebox{\columnwidth}{!}{%
\begin{tabular}{l|cc|cc|c}
\toprule
\textbf{Method} & \textbf{CLIP-I}↑ & \textbf{DINO-I}↑ & \textbf{CLIP-Score}↑ & \textbf{Overall cons.}↑ & \textbf{Aesthetic}↑ \\
\midrule
$\tau=0.2$ & 0.6465 & 0.4547 & 0.3098 & 0.2600 & 0.6200 \\
\textbf{Ours(\textbf{$\tau=0.5$})} & \textbf{0.6504} & 0.4566 & 0.3107 & 0.2691 & \textbf{0.6227} \\
$\tau=1.0$ & 0.6482 & \textbf{0.4568} & \textbf{0.3115} & \textbf{0.2713} & 0.6154 \\
\bottomrule
\end{tabular}
}
\caption{
\textbf{Ablation on the injection timestep 
$\tau$
}. Experimental results show that a value of 
$\tau=0.5$
 strikes the best balance between high edit fidelity and strong structural coherence.}
\label{tab:ablation_tau}
\end{table}

\section{Conclusion}
%We present ContextFlow, a training-free framework for video object editing. Our core contribution, Adaptive Context Enrichment, injects controlled structural guidance from a parallel reconstruction path via Key-Value concatenation mechanism. By targeting vital layers identified through Guidance Responsiveness analysis, our method balances edit fidelity with background stability. Experiments confirm the effectiveness of our approach in creating high-quality, consistent video object edits, empowering users with fine-grained creative control.

We present ContextFlow, a training-free framework specifically designed for addressing the key challenges of video object editing, including preserving temporal continuity and maintaining object-background consistency. Our core contribution, Adaptive Context Enrichment, strategically injects controlled structural guidance that is extracted and refined from a dedicated parallel reconstruction path. This injection process relies on a custom-built Key-Value concatenation mechanism, which ensures smooth information integration.

By precisely targeting vital layers that are identified through in-depth Guidance Responsiveness analysis, our method effectively balances edit fidelity with background stability. Experiments conducted across diverse test scenarios confirm the effectiveness of our approach in creating high-quality, temporally consistent video object edits, ultimately empowering users with intuitive fine-grained creative control over the editing process.
\clearpage
\setcounter{page}{1}
\maketitlesupplementary

This supplementary document provides additional details, experiments, and visualizations to complement our main paper. The contents are organized as follows:
\begin{itemize}
\item \textbf{Implementation and Evaluation Details.} Expanding on the main text, we provide a comprehensive overview of our experimental setup, computational cost, evaluation protocol specifics, and a deeper dive into our Guidance Responsiveness analysis.
\item \textbf{Additional Qualitative Comparisons.} We present more visual comparisons against some state-of-the-art video editing techniques methods, specifically Unic~\cite{ye2025unic} and Pika~\cite{PikaLabs2025Pika2}, to further highlight the superiority of ContextFlow.
\item \textbf{Visual Analysis for Ablation Studies.} We provide visual evidence to support the quantitative results of our ablation studies presented in the main paper.
\item \textbf{Limitations and Future Work.} We discuss the current limitations of our method and suggest potential directions for future research.
\end{itemize}

\section{Implementation and Evaluation Details}
\label{sect:A}

\subsection{Experimental Setup and Computational Cost}
This section expands on the implementation details mentioned in the main paper. Our framework is implemented in Python using PyTorch. All experiments were conducted on a server with two NVIDIA A800 (80GB) GPUs. We build our training-free framework upon the publicly available Wan2.1-I2V-14B-480P~\cite{wan2025wan} model. The dual-path generation process, which is central to our \textbf{Adaptive Context Enrichment} mechanism, requires approximately 120GB of VRAM in total (~60GB per GPU). For a typical 81-frame video at 480p resolution, the dual-path generation process takes approximately 25 minutes. As our method propagates an edit from the first frame, we utilize off-the-shelf image editing tools to generate the initial edited frame $I_{edit}$. For the experiments presented, we use AnyDoor~\cite{chen2024anydoor} for object insertion, InsertAnything~\cite{song2025insert} for object swapping, and MagicQuill~\cite{liu2025magicquill} for object deletion.

\subsection{Further Details on Guidance Responsiveness Analysis}
Expanding on the Guidance Responsiveness analysis presented in Figure 5 of the main paper, we offer details on its robustness. The $GR$ curves were derived by averaging results from a diverse set of 10 videos for each task, including in-domain videos from the Unic-Benchmark~\cite{ye2025unic} and out-of-domain videos from sources like the DAVIS dataset~\cite{ponttuset20182017davischallengevideo}.

\subsection{Evaluation Protocol Details}
We provide further clarification on the metrics used for quantitative evaluation.

\paragraph{Identity Preservation (CLIP-I / DINO-I).} To measure how well the identity of the reference object is preserved, we compute frame-wise similarity. For CLIP-I, we use the pre-trained clip-vit-large-patch14~\cite{Radford2021Learning} model from OpenAI. For DINO-I, we use the dinov2-giant~\cite{oquab2023dinov2} model from Facebook. The process is as follows: (1) we extract the image feature of the reference object; (2) for each frame in the generated video, we extract its image feature; (3) we compute the cosine similarity between the reference feature and each frame’s feature; (4) the final score is the average similarity across all frames. 

\paragraph{Background Preservation (PSNR / SSIM).} These metrics are computed exclusively on the unedited regions of the video to measure background stability. The unedited area is defined by a mask. For object swapping and deletion tasks on the Unic-Benchmark, we use the ground-truth masks provided with the dataset. For the object insertion task, where no prior object exists, we generate a mask for the newly inserted object using Grounding-SAM~\cite{ren2024grounded}, and the background is defined as the area outside this mask. The final PSNR and SSIM scores are averaged over all frames.

\paragraph{VBench Metrics.} For the reader’s convenience, we briefly summarize the calculation methods for the VBench metrics used in our paper, drawing from the descriptions in the official VBench suite~\cite{huang2024vbench}.

\begin{itemize}
\item[\textbullet] \textbf{Overall Consistency.} This metric evaluates the overall video-text consistency, reflecting both semantic and style alignment. It is computed using ViCLIP~\cite{Wang2023InternVid} as an aiding metric, where the model assesses the video against text prompts that contain different semantics and styles.
\item[\textbullet] \textbf{Motion Smoothness.} To evaluate temporal quality beyond simple appearance consistency, this metric assesses whether the motion in the generated video is smooth and follows the physical laws of the real world. It utilizes the motion priors from a video frame interpolation model~\cite{Li2023AMT} to quantify the smoothness of the generated motions.

\item[\textbullet] \textbf{Dynamic Degree.} Complementing smoothness, this metric measures the degree of dynamics (i.e., the presence of large motions) to ensure that completely static videos are not unfairly favored in temporal quality scores. It uses the RAFT~\cite{Teed2020RAFT} model to estimate the degree of dynamics in synthesized videos.

\item[\textbullet] \textbf{Aesthetic Quality.} This metric evaluates the frame-wise artistic and beauty value perceived by humans. It employs the LAION~\cite{LAION-AI} aesthetic predictor to assess aspects such as the layout, the richness and harmony of colors, photo-realism, naturalness, and the overall artistic quality of the video frames.
\end{itemize}

\section{Additional Qualitative Comparisons}
\begin{figure*}[h!]
\centering
\includegraphics[width=1.0\textwidth]{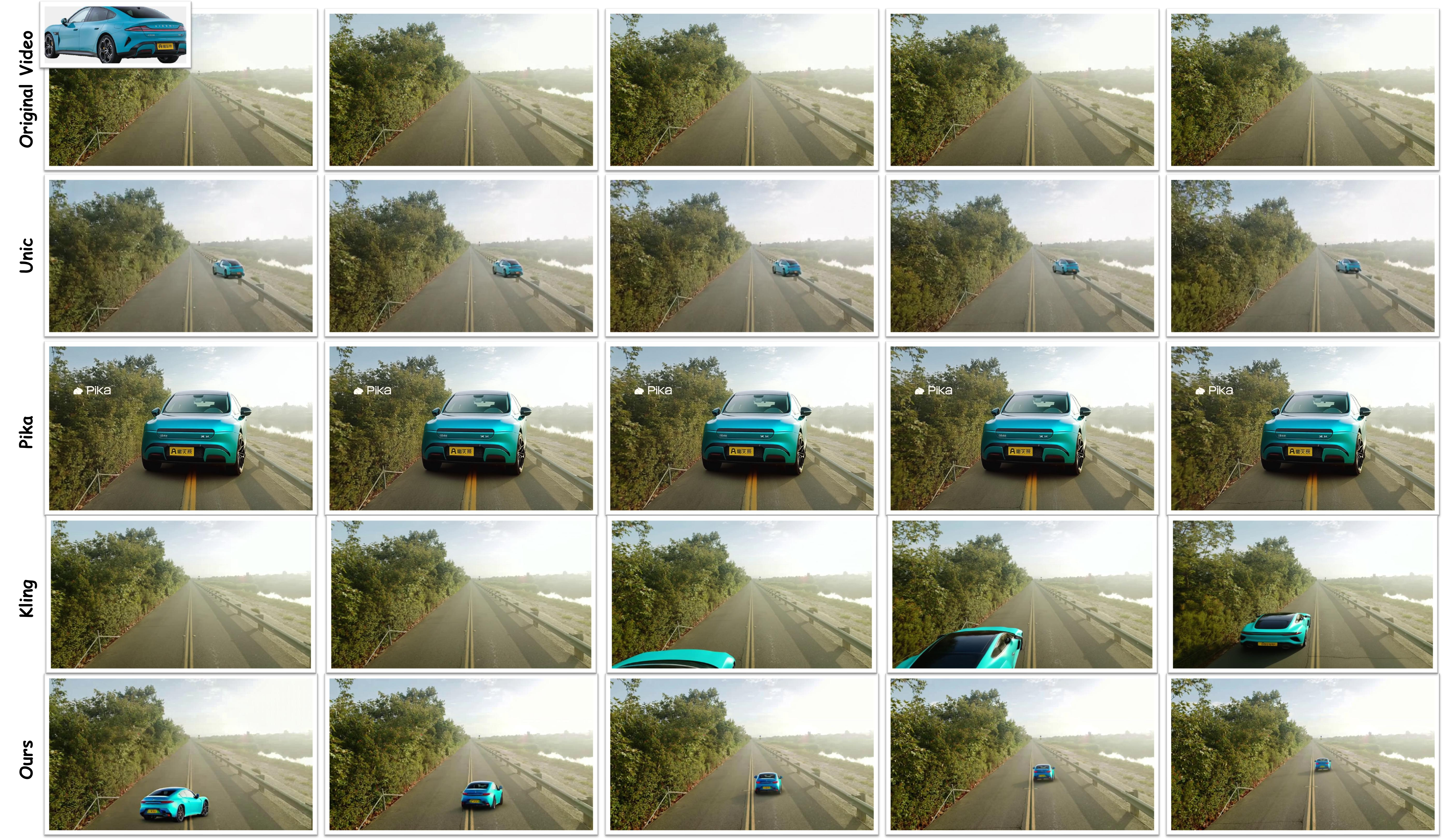}
\caption{\textbf{Visualization of Qualitative Comparison.} Target prompt: ``A blue car is driving on the road.'' Our method accurately simulates the movement state of the blue car traveling on the road, whereas the other methods fail to achieve the effect of the car moving correctly in the proper position on the road.}
\label{1}
\end{figure*}

\begin{figure*}[h]
\centering
\includegraphics[width=1.0\textwidth]{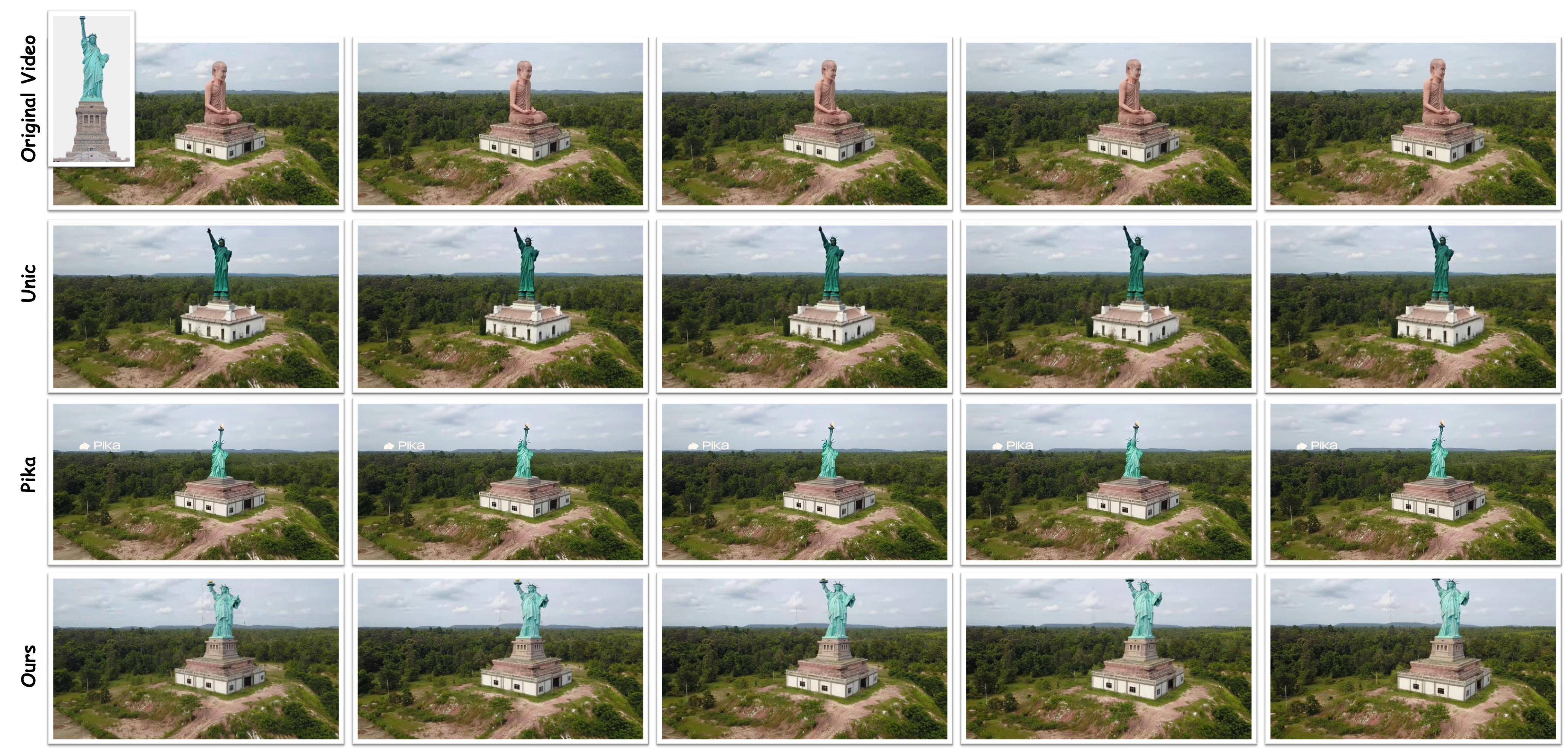}
\caption{\textbf{Visualization of Qualitative Comparison.} Target prompt: ``A colossal Statue of Liberty stands atop a white base with classical columns, surrounded by a dense forest.'' Our method achieves the highest fidelity to the reference image.}
\label{2}
\end{figure*}

\begin{figure*}[h]
\centering
\includegraphics[width=1.0\textwidth]{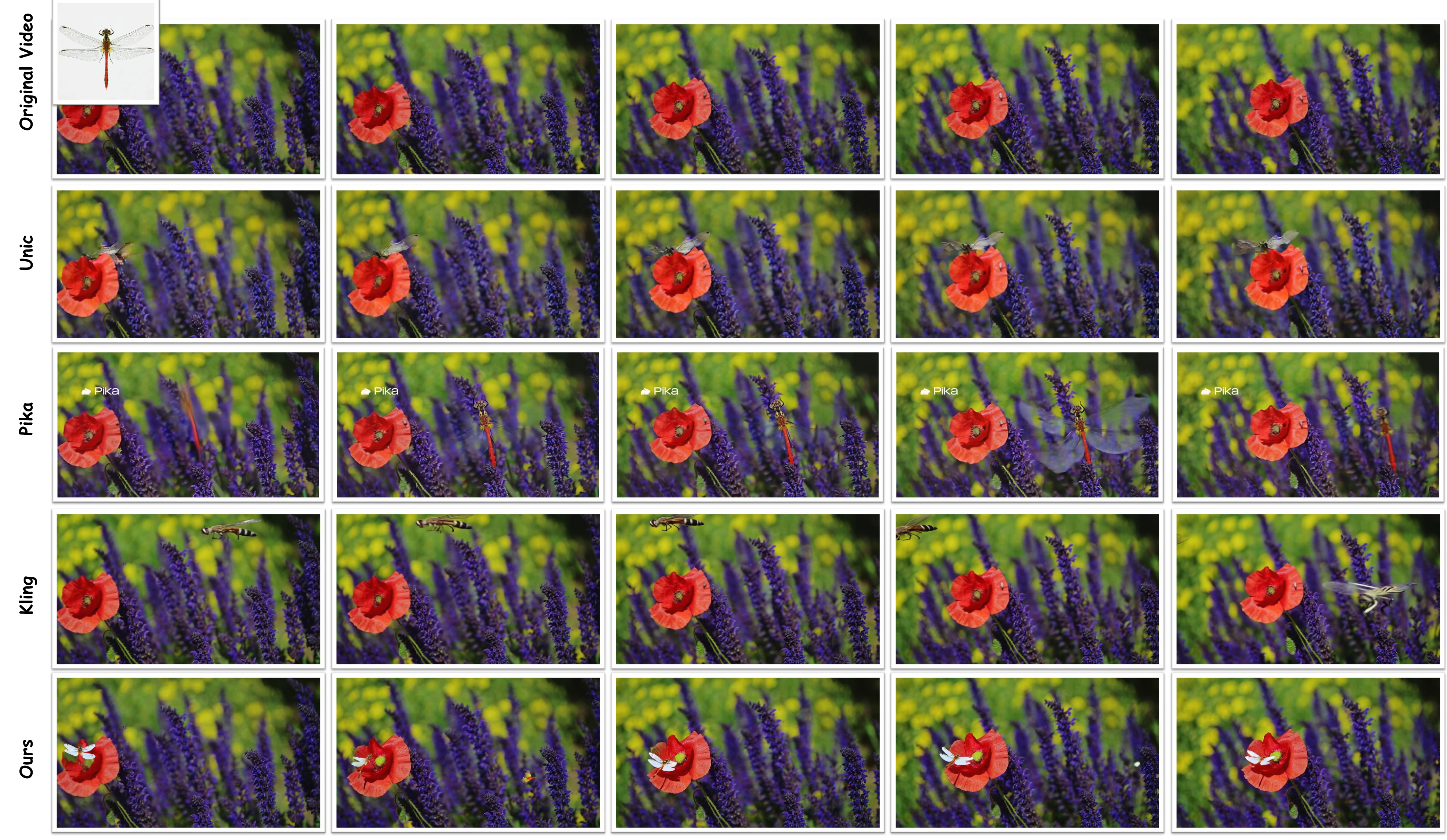}
\caption{\textbf{Visualization of Qualitative Comparison.} Target prompt: ``A red poppy flower surrounded by purple flowers. A dragonfly is stopping at the red poppy flower.'' Our method has achieved results that most closely to match the reference image as well as the target prompt (i.e., a dragonfly is ``stopping at the red poppy flower''). }
\label{3}
\end{figure*}

\begin{figure*}[h!]
\centering
\includegraphics[width=1.0\textwidth]{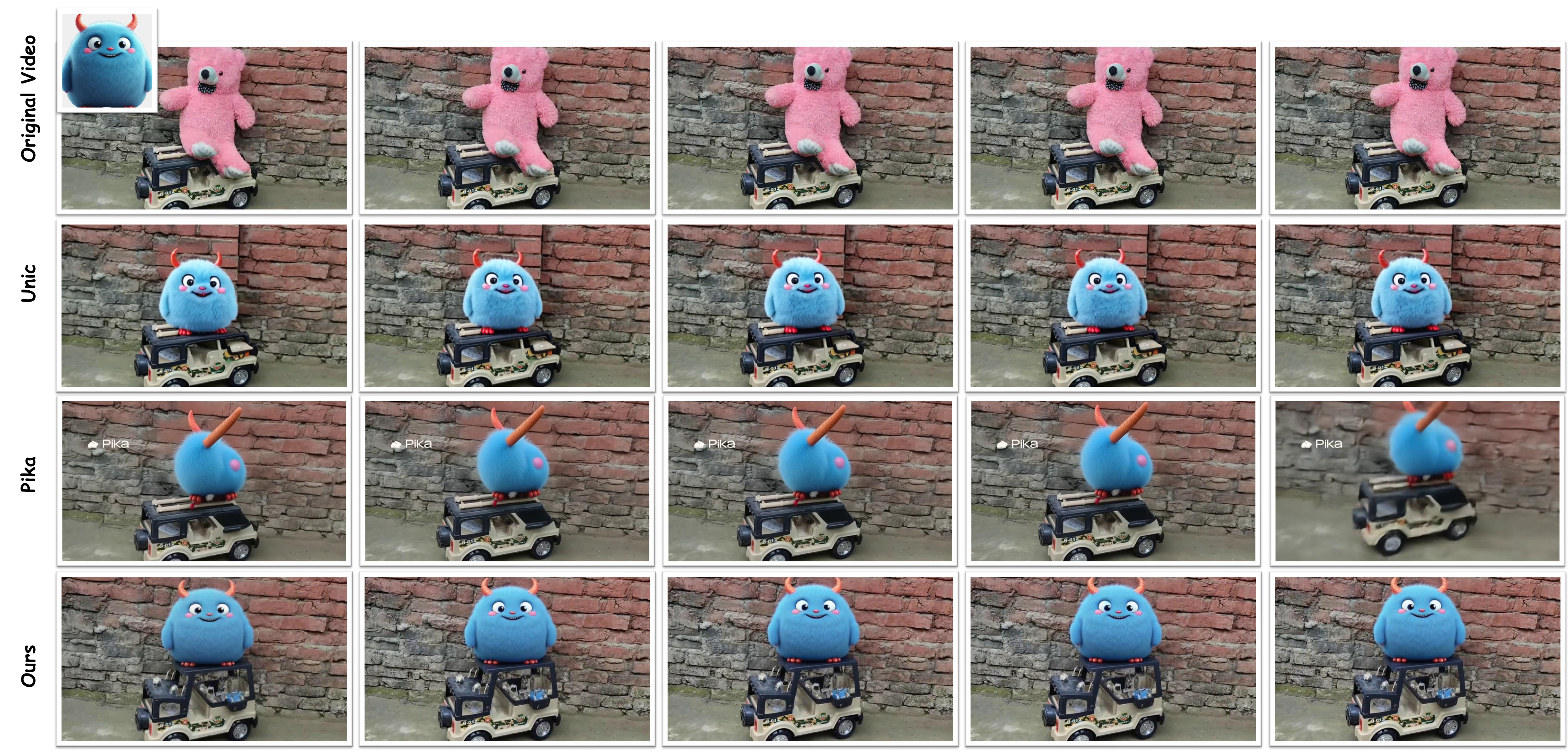}
\caption{\textbf{Visualization of Qualitative Comparison.} Target prompt: ``A cute blue fluffy monster sits on top of a toy jeep against a brick wall.'' Our method achieves the highest fidelity to the reference image, and it has also achieves excellent preservation of the shape of the brick walls in the original video.}
\label{14}
\end{figure*}

\begin{figure*}[h!]
\centering
\includegraphics[width=1.0\textwidth]{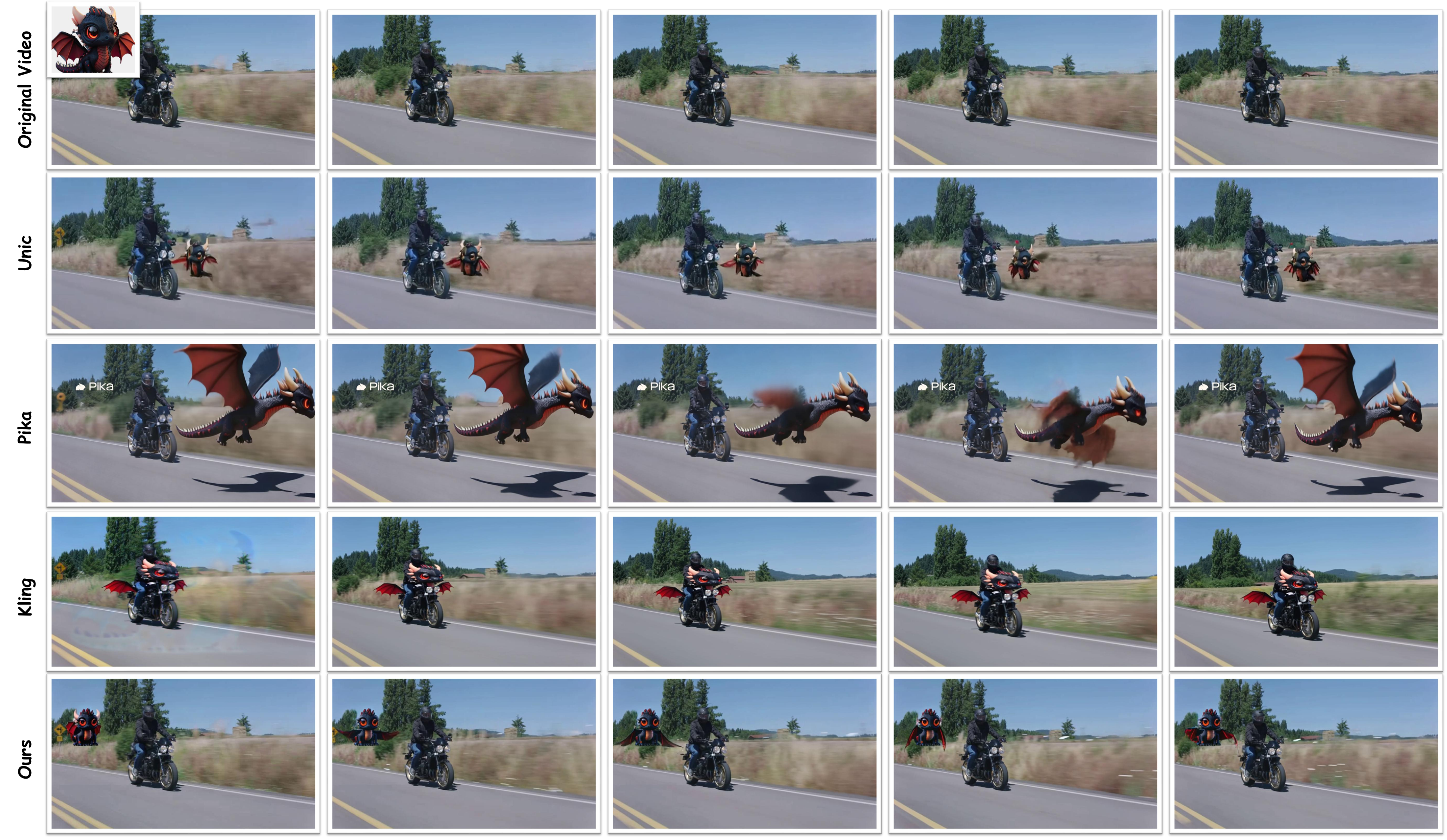}
\caption{\textbf{Visualization of Qualitative Comparison.} Target prompt: ``A man rides a motorcycle down a lonely road, with a little flying dragon. The flying dragon is near the man through the video. It has the same speed with the man.'' Our method achieves the highest fidelity to the reference image, while the movements of the little dragon are also the most natural.}
\label{4}
\end{figure*}

\begin{figure*}[h!]
\centering
\includegraphics[width=1.0\textwidth]{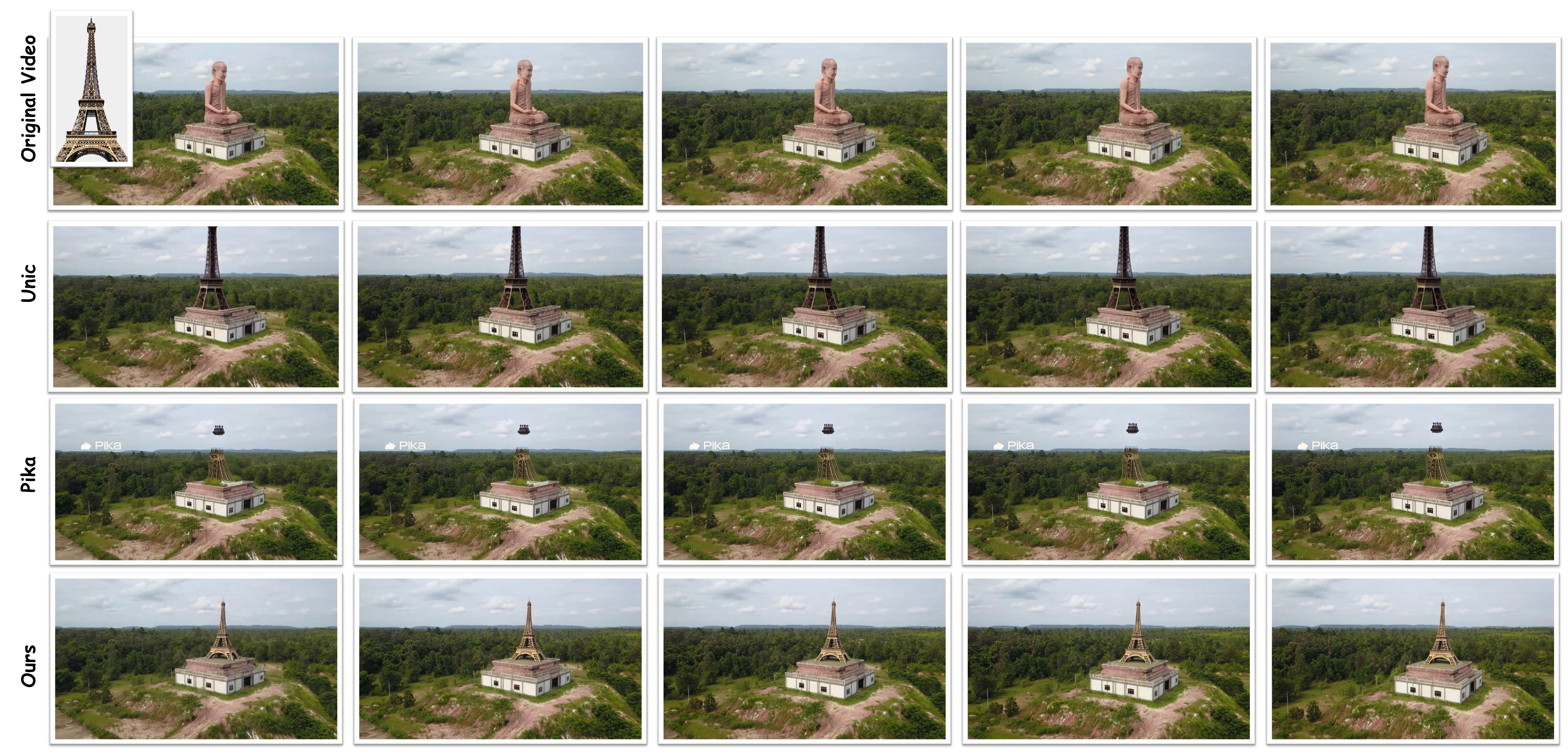}
\caption{\textbf{Visualization of Qualitative Comparison.} Target prompt: ``A colossal Eiffel Tower stands atop a white base with classical columns, surrounded by a dense forest.'' Our method achieves the highest fidelity to the reference image.}
\label{18}
\end{figure*}

\begin{figure*}[h!]
\centering
\includegraphics[width=1.0\textwidth]{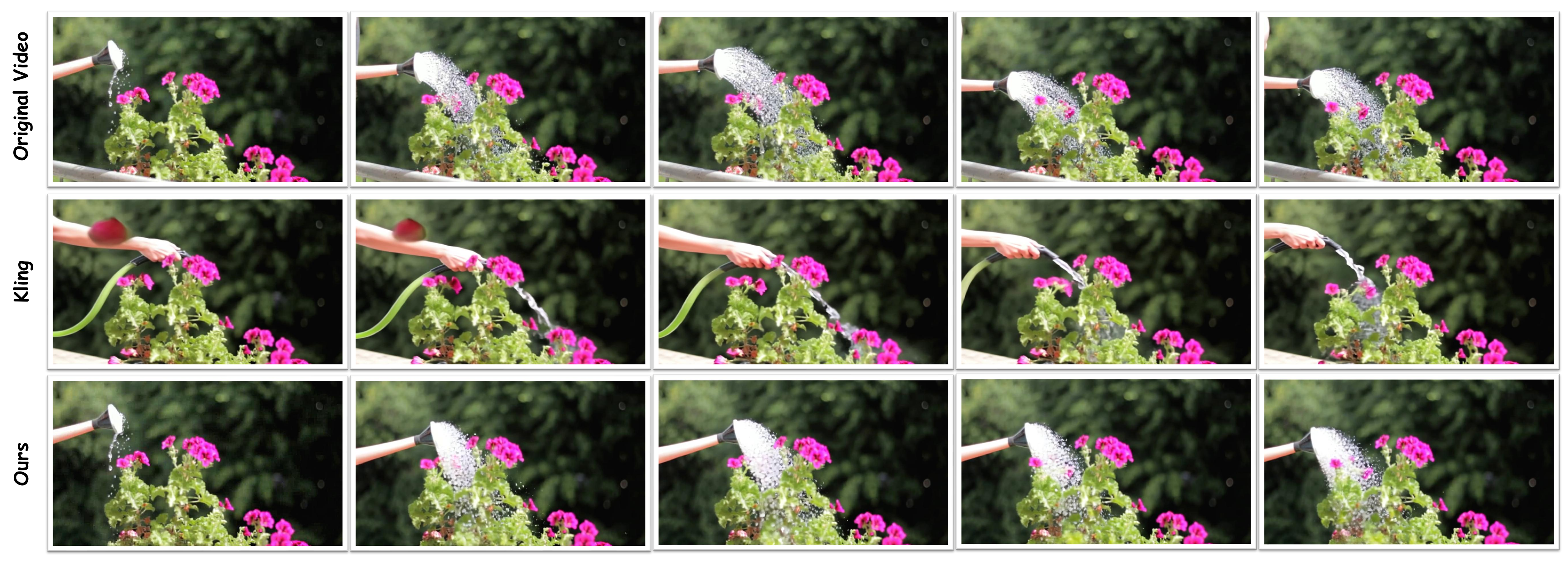}
\caption{\textbf{Visualization of Qualitative Comparison.} Target prompt: ``A hand-held white plastic watering can is used to water a vibrant collection of pink and purple flowers, including geraniums and possibly petunias, in a white planter.'' The target to be deleted in this video is the railing in the lower left corner. Our method has high fidelity for the unedited areas, and its effect is better than that of Kling ~\cite{KlingAI_Global}.}
\label{17}
\end{figure*}

\begin{figure*}[h!]
\centering
\includegraphics[width=1.0\textwidth]{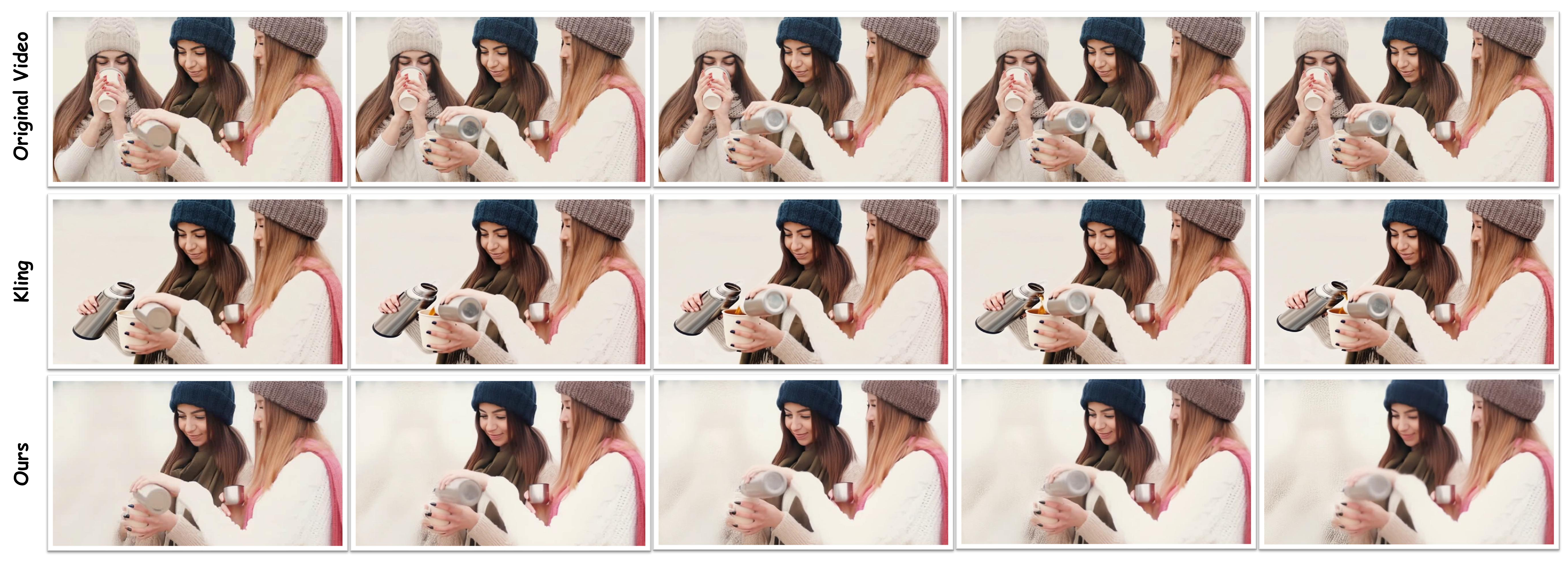}
\caption{\textbf{Visualization of Qualitative Comparison.} Target prompt: ``Two young women, dressed in cozy winter attire, stand together in a tranquil snowy winter landscape, engaging in the warmth of sharing a hot drink.'' The target to be deleted in this video is the girl on the far left. Our method performs better.}
\label{16}
\end{figure*}

To further situate ContextFlow’s performance, we provide qualitative comparisons against more strong, contemporary methods, for example Unic~\cite{ye2025unic} and commercial training-based tools like Pika ~\cite{PikaLabs2025Pika2} and Kling ~\cite{KlingAI_Global}. As shown in Figure \ref{1} - Figure \ref{16} , ContextFlow consistently demonstrates superior performance in both identity preservation and motion consistency.

ContextFlow accurately maintains the intricate details and colors of the reference image, while ensuring it interacts realistically with the clouds. Both Unic and Pika also allow users to input reference images to define objects for insertion or swapping. However, they cannot enable users to freely and customarily edit the first frame, a limitation that causes them to struggle to maintain such high fidelity to the reference image.

\section{Visual Analysis for Ablation Studies}

To complement the quantitative ablation studies in the main paper, this section provides corresponding visual analysis. These visualizations offer intuitive insights into how each component of ContextFlow contributes to the final result.

\subsection{Visualization of Ablation on Guidance Strategy (Table \ref{tab:ablation_strategy}).} 
\begin{figure*}[h]
\centering
\includegraphics[width=1.0\textwidth]{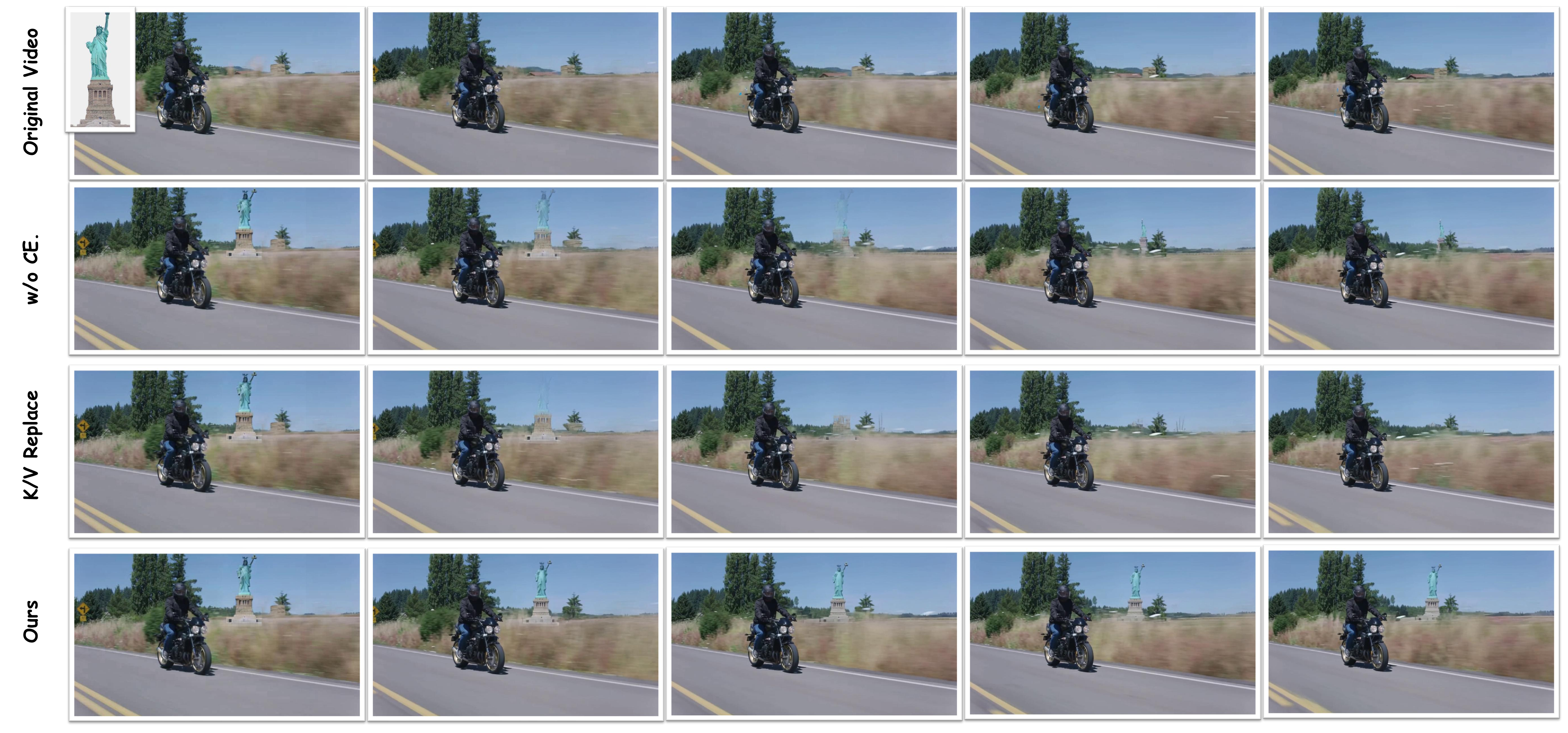}
\caption{\textbf{Visualization of Ablation on Guidance Strategy.} Target prompt: ``A man rides a motorcycle down a lonely road. A large, imposing Statue of Liberty stands in the background.''  Neither Method ``w/o CE.'' nor Method ``K/V replacement'' can effectively integrate the Statue of Liberty into the background correctly and naturally, resulting in the inserted object either disappearing or failing to blend well into the background environment. Please zoom in for a closer check.}
\label{5}
\end{figure*}

\begin{figure*}[h]
\centering
\includegraphics[width=1.0\textwidth]{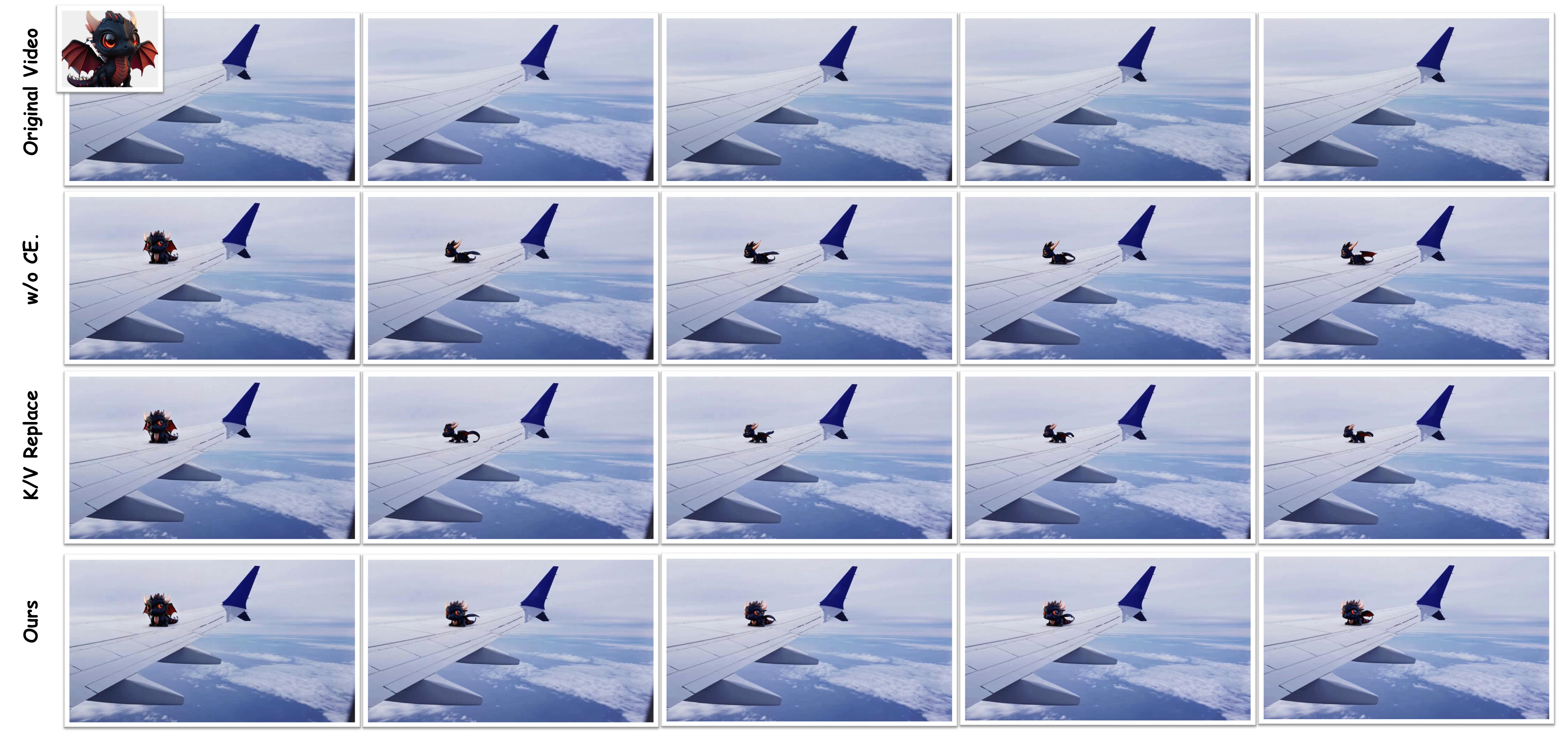}
\caption{\textbf{Visualization of Ablation on Guidance Strategy.} Target prompt: ``A cute, black and red dragon with large, glowing eyes stands on the plane wing, playfully flipping its wings.''  Both Method ``w/o CE.'' and  Method ``K/V replacement'' struggle to maintain the little dragon's shape in conformity with the reference image as it was in the first frame; instead, strange deformations occur. Please zoom in for a closer check.}
\label{6}
\end{figure*}

\begin{figure*}[h]
\centering
\includegraphics[width=1.0\textwidth]{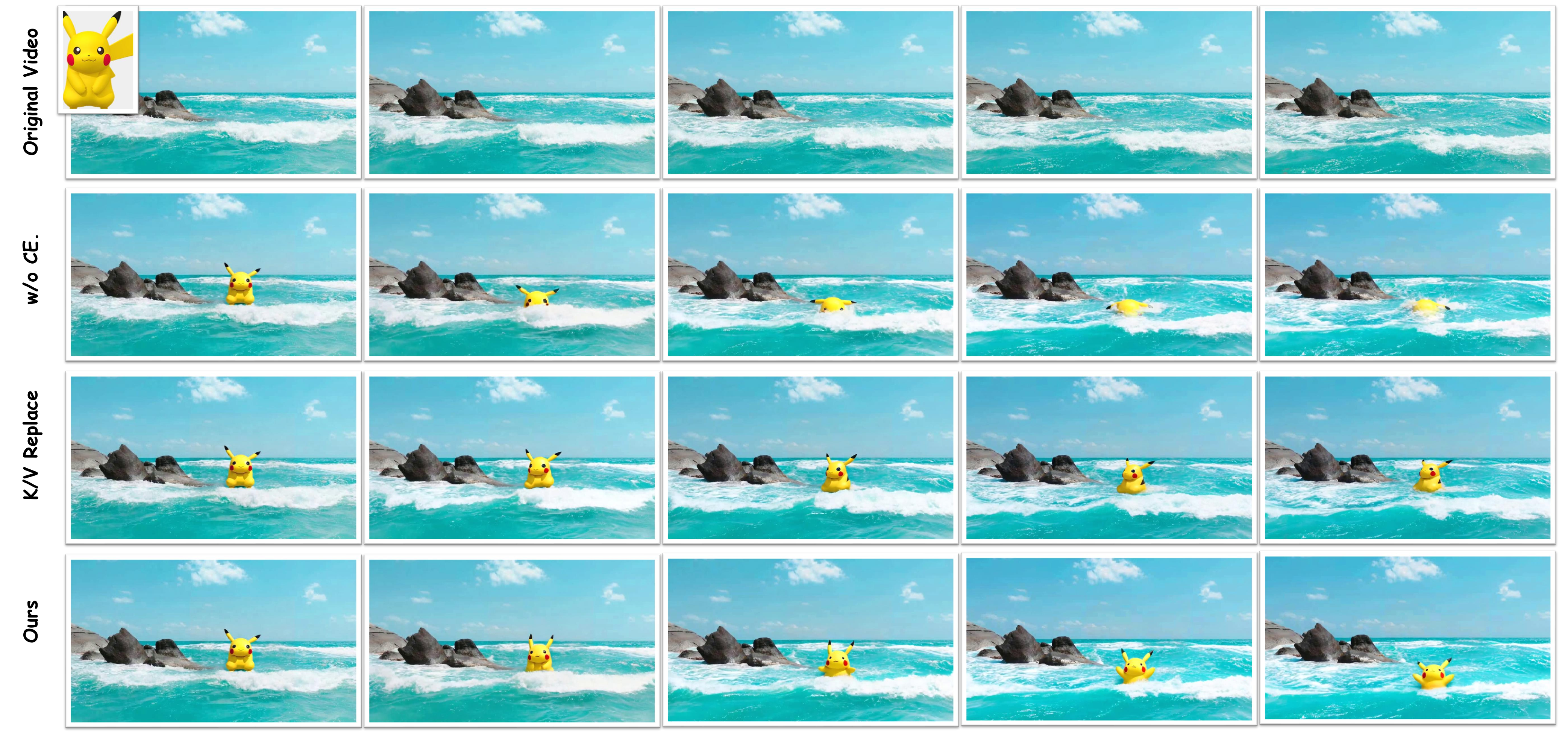}
\caption{\textbf{Visualization of Ablation on Guidance Strategy.} Target prompt: ``A Pikachu is floating on the sea.'' Method ``w/o CE.'' fails to properly integrate Pikachu with the background, thus failing to achieve the effect required by the prompt---``floating on the sea surface''. In contrast, Pikachu in Method ``K/V replacement'' has extremely stiff movements and lacks the natural sense of dynamics associated with floating. Please zoom in for a closer check.}
\label{7}
\end{figure*}

Figure \ref{5} - Figure \ref{7} visualizes the impact of our core guidance mechanism. Without any adaptive Context Enrichment (CE), the inserted object struggles to maintain a consistent identity and motion. Using “K/V Replacement” instead of our concatenation approach leads to severe artifacts and identity degradation, validating our hypothesis that hard replacement causes destructive contextual conflict. Our full method successfully fuses the object while preserving the background.

\subsection{Ablation on Amount of Guidance (Table \ref{tab:ablation_k}).} 
\begin{figure*}[h]
\centering
\includegraphics[width=0.8\textwidth]{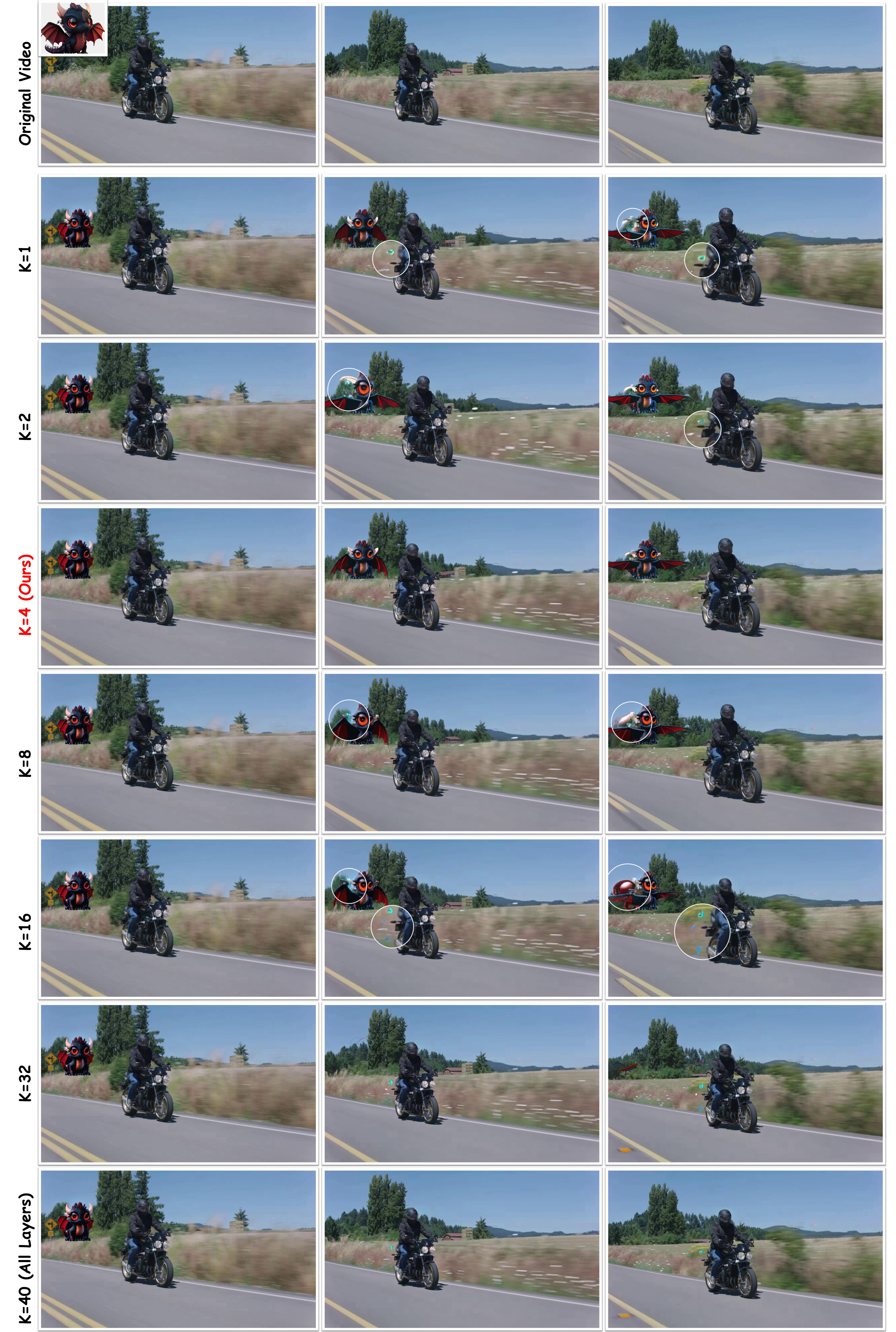}
\caption{\textbf{Visualization of Ablation on Amount of Guidance.} Target prompt: ``A man rides a motorcycle down a lonely road, with a little flying dragon. The flying dragon is near the man through the video. It has the same speed with the man.'' When \( k \leq 2 \) or \( 8 \leq k \leq 16 \), the generated video exhibits strange colored bubbles and abnormal deformation of the little dragon's ears, leading to poor editing quality (enlarged via ``magnifying glass'' in the figure). When \( k \geq 32 \) the little dragon disappears. Please zoom in for a closer check.}
\label{8}
\end{figure*}

\begin{figure*}[h]
\centering
\includegraphics[width=1.0\textwidth]{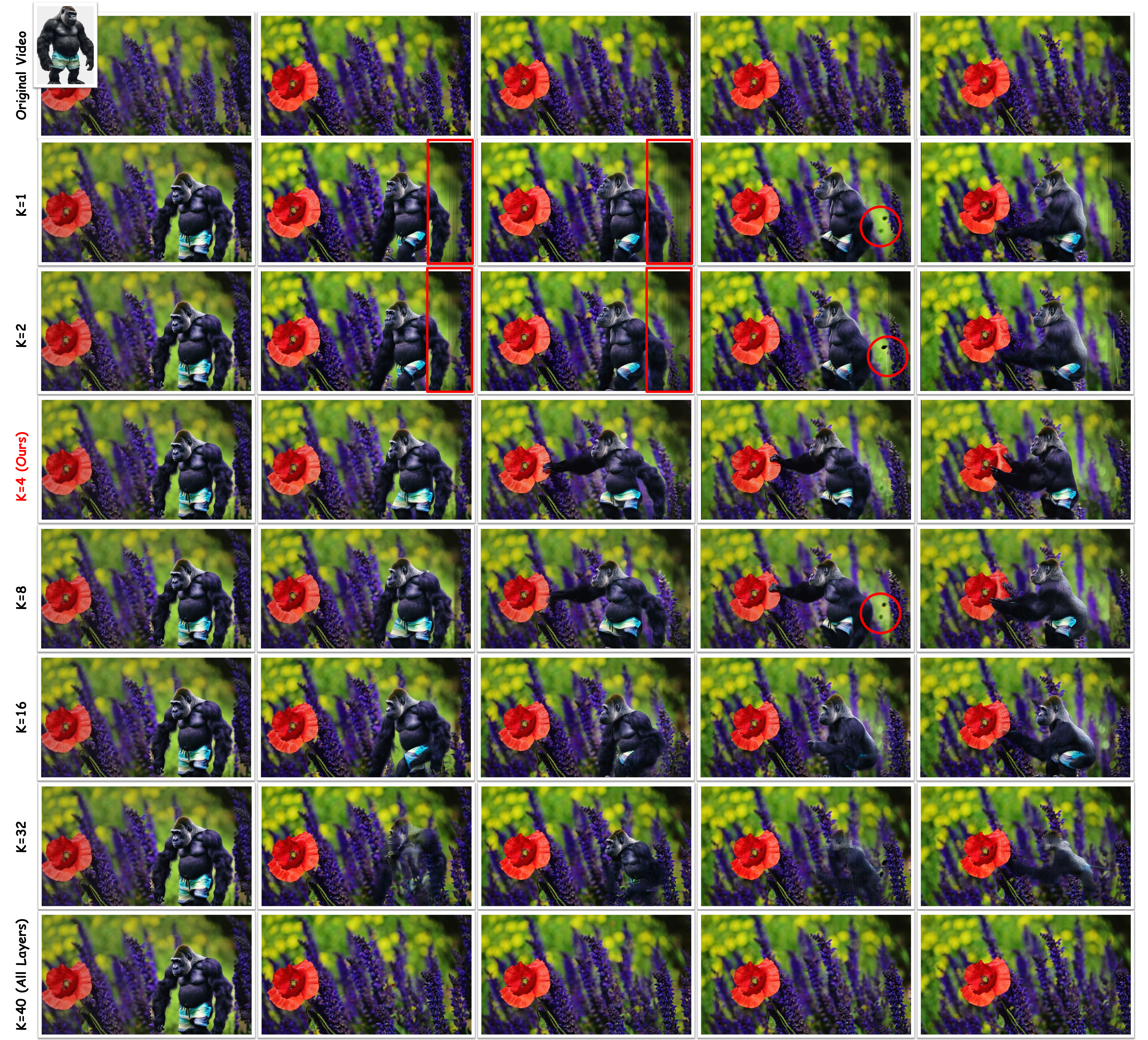}
\caption{\textbf{Visualization of Ablation on Amount of Guidance.} Target prompt: ``A red poppy flower surrounded by purple flowers. A large gorilla is gently trying to touch the red poppy flower.'' When \( k \leq 2 \), the generated video will have discordant black stripes (zoom in and see the red box area in the figure); when \( k \geq 8 \), the movement posture of the gorilla will appear strange; when \( k \geq 32 \), the gorilla will become translucent or disappear in subsequent frames. }
\label{9}
\end{figure*}

Figure \ref{8} and Figure \ref{9} shows the effect of varying $k$, the number of guided layers. With insufficient guidance ($k=1$), the object appears unstable, which causes poor video quality. With excessive guidance ($k=40$, all layers), the object becomes overly constrained by the background context. Our choice of $k=4$ strikes an optimal balance, yielding a dynamic yet stable object.

\subsection{Ablation on Layer Selection (Table \ref{tab:ablation_selection}).} 
\begin{figure*}[h]
\centering
\includegraphics[width=1.0\textwidth]{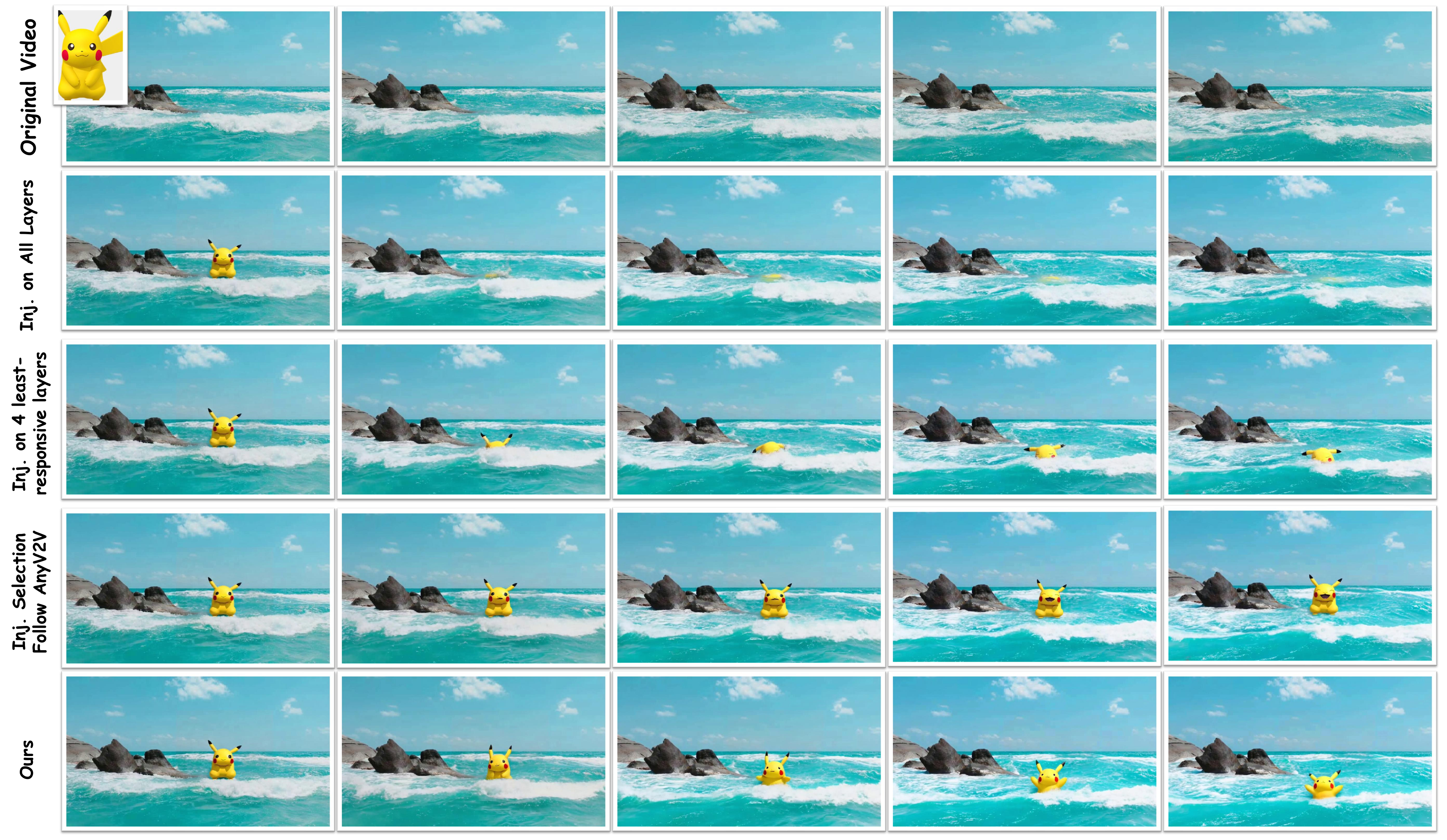}
\caption{\textbf{Visualization of Ablation on Layer Selection.} Target prompt: ``A Pikachu is floating on the sea.'' Injection on all layers and Injection on 4 least responsive layers will cause Pikachu to disappear quickly in subsequent frames or fail to display actions correctly according to the target prompt. Injection selection follow AnyV2V~\cite{ku2024anyv2v} will produce an extremely stiff and weird ``floating'' effect that is highly unnatural, along with an odd smile not required by the prompt. Please zoom in for a closer check.}
\label{10}
\end{figure*}

\begin{figure*}[h]
\centering
\includegraphics[width=1.0\textwidth]{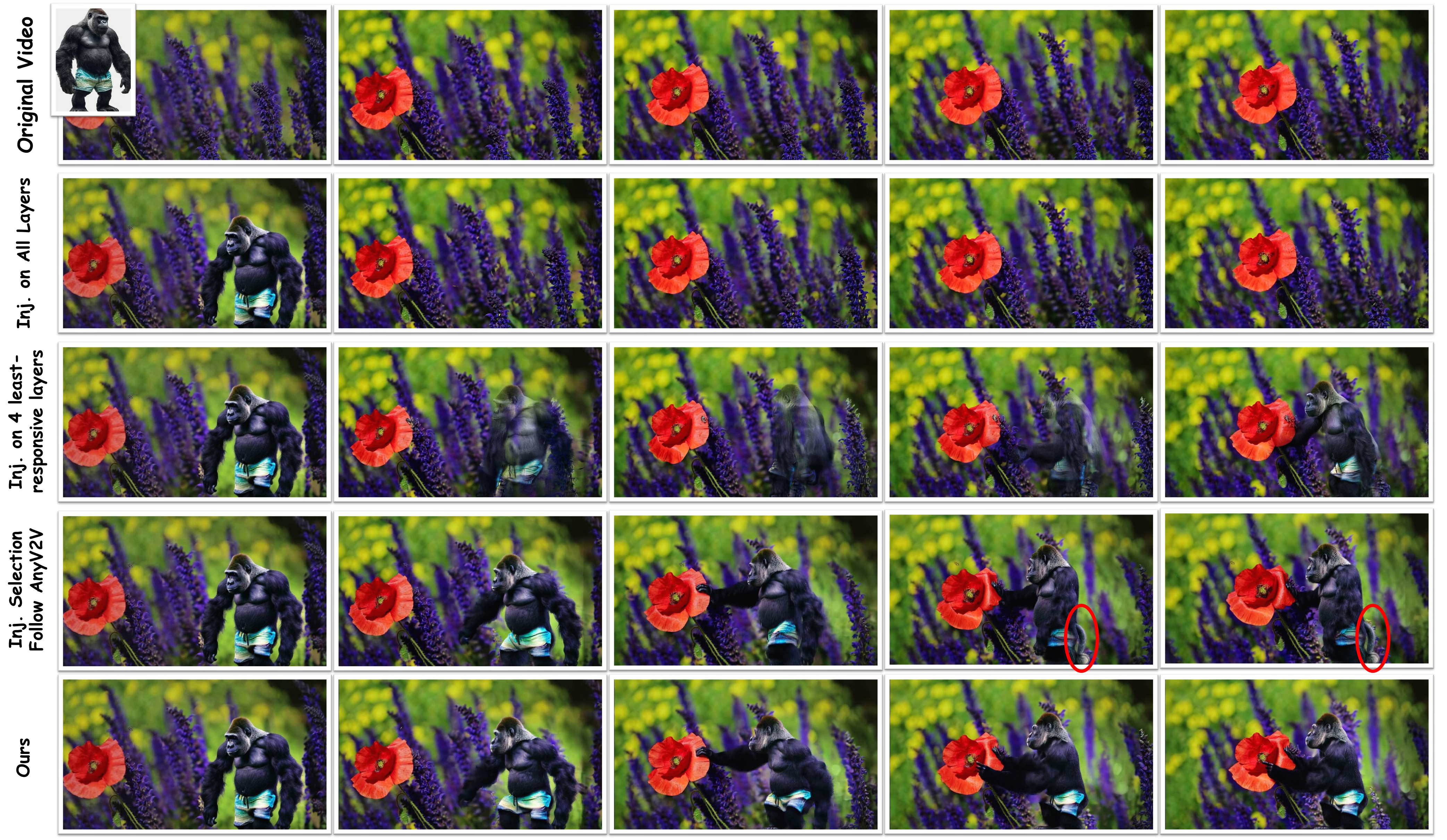}
\caption{\textbf{Visualization of Ablation on Layer Selection.} Target prompt: ``A red poppy flower surrounded by purple flowers. A large gorilla is gently trying to touch the red poppy flower.'' Injection on all layers and Injection on 4 least responsive layers will cause the inserted object to disappear or become translucent. Injection selection follow AnyV2V~\cite{ku2024anyv2v} will once again  result in the strange form of the gorilla: an oddly appearing tail (within the red circle). Please zoom in for a closer check.}
\label{11}
\end{figure*}

Figure \ref{10} and Figure \ref{11} validates our data-driven layer selection strategy. Injecting information into all layers will lead to an excessive amount of information from the reconstruction path being introduced into the editing path, making it difficult to maintain the edited objects and causing the generated video to unconsciously lean towards the original video. Injecting guidance into the four least-responsive layers yields poor results, similar to providing weak guidance. Using a U-Net-based heuristic (from AnyV2V~\cite{ku2024anyv2v}) in the DiT model, which in a sense amounts to randomly selecting certain layers within DiT for injection, still leads to suboptimal results. Due to the inability of the injected layers to solve substantive problems, it often leads to stiff interaction between new objects and the background, and even unconscious emergence of unreasonable phenomena (such as the chimpanzee in Figure \ref{11} strangely growing a tail). This visually confirms that our Guidance Responsiveness metric correctly identifies the most impactful layers for intervention.

\subsection{Ablation on Guidance Window (Table \ref{tab:ablation_tau}).} 
\begin{figure*}[h]
\centering
\includegraphics[width=1.0\textwidth]{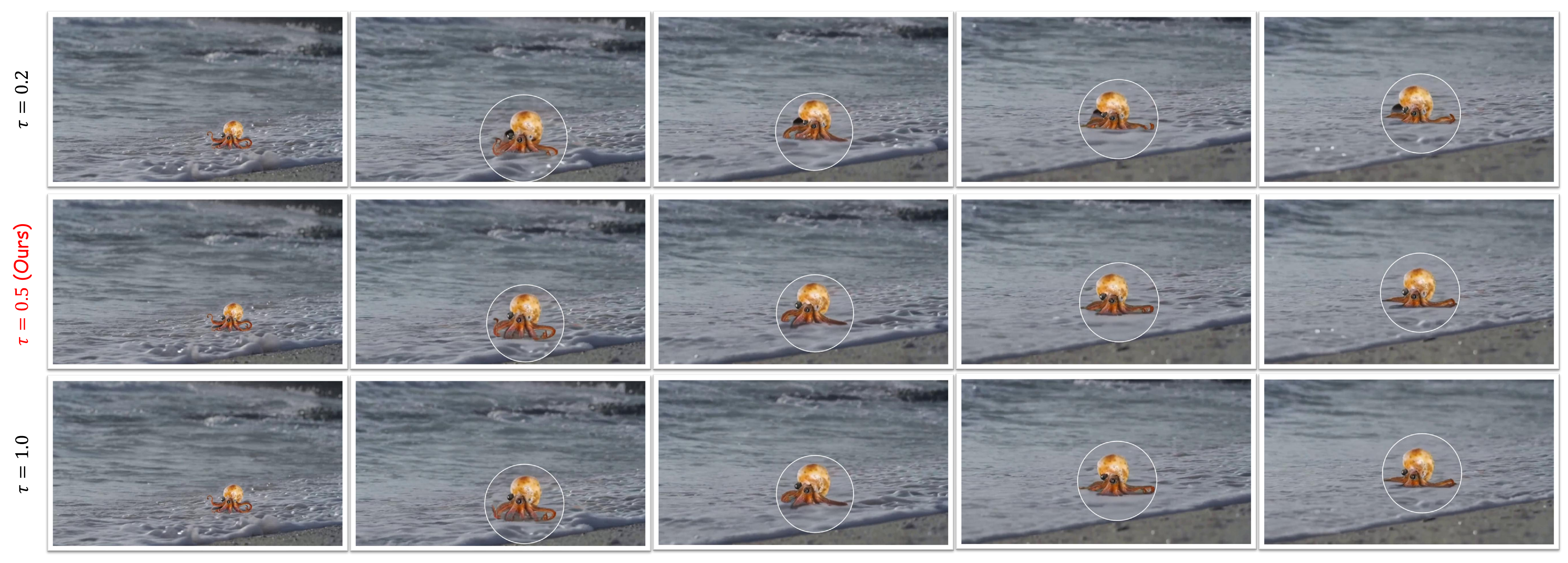}
\caption{\textbf{Visualization of Ablation on Guidance Window.} Target prompt: ``An octopus is swimming in the sea.'' A magnifying effect is applied to zoom in on the octopus in the image. $\tau=0.2$ causes the octopus to take on an abnormal shape with odd black patches, making it unable to blend well with the environment. In contrast, under $\tau=0.5$ and $\tau=1.0$, the octopus has a more natural shape and blends appropriately with the environment. Please zoom in for a closer check.}
\label{12}
\end{figure*}

\begin{figure*}[h!]
\centering
\includegraphics[width=1.0\textwidth]{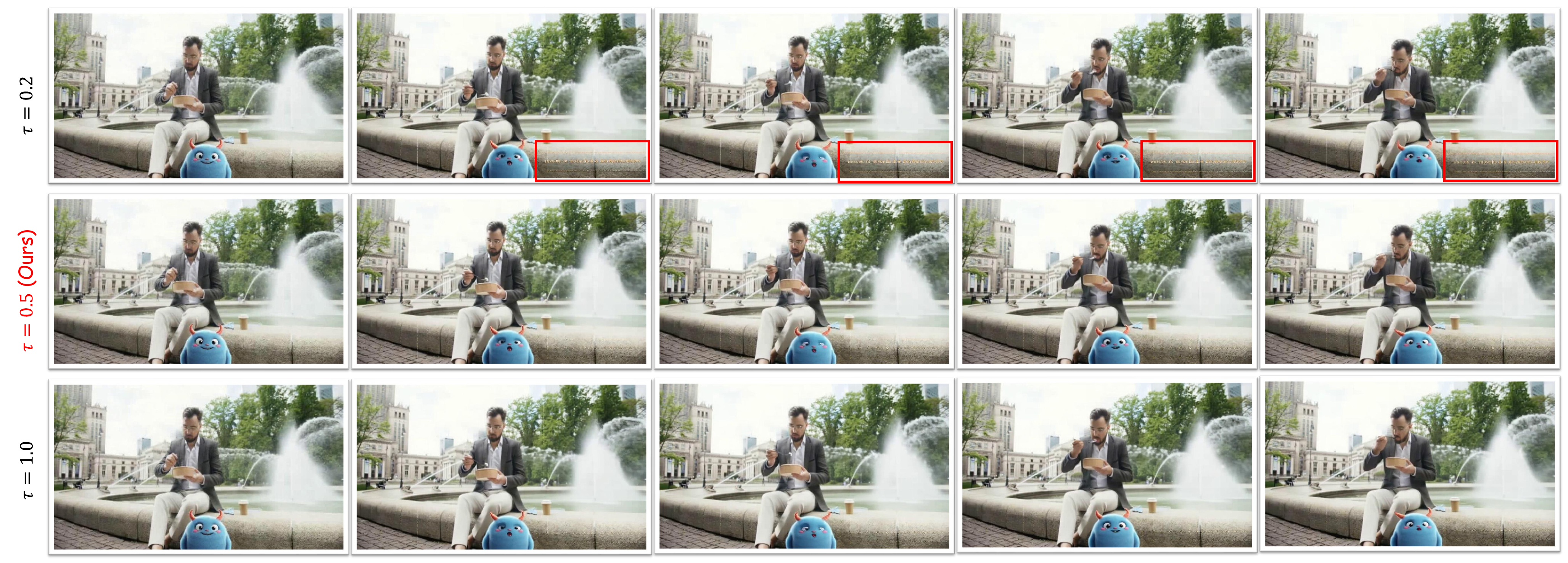}
\caption{\textbf{Visualization of Ablation on Guidance Window.} Target prompt: ``A man in a dark grey blazer and white shirt is seated on a city fountain's edge, eating from a wooden bowl with a fork. A cute blue fluffy monster is on the ground.'' $\tau=0.2$ will cause an abnormal text-like pattern to appear in the red box at the bottom right corner of the picture, while such a pattern will not appear under $\tau=0.5$ and $\tau=1.0$. Please zoom in for a closer check.}
\label{13}
\end{figure*}

Figure \ref{12} and Figure \ref{13} illustrates the importance of the timestep threshold $\tau$. A value of $\tau=0.2$ often results in insufficient fusion between the edited object and its background, leading to abnormal deformations of the edited object or unnatural modifications to the surrounding background. In contrast, $\tau=0.5$ effectively addresses these issues, achieving both excellent fusion quality and robust preservation of the reference identity. Notably, the performance with $\tau=1.0$ is comparable to that of $\tau=0.5$. This observation suggests that through the fusion of the editing path with the reconstruction path via the \textbf{Adaptive Context Enrichment} mechanism during the first half of the diffusion process, sufficient essential information has already been acquired. Consequently, during the latter half of the diffusion process, there is minimal need to further extract useful information from the reconstruction path. Considering the trade-off between computational efficiency and editing performance, $\tau=0.5$ emerges as the optimal choice.

\clearpage

\clearpage
\section{Limitations and Future Work}

Despite its strong performance, ContextFlow has several limitations that open avenues for future research.

\paragraph{Dependency on First-Frame Edit.} The quality of our video edits is inherently tied to the quality of the initial edited frame, $I_{edit}$ . Artifacts or inaccuracies in the first-frame edit, produced by third-party image editors, are likely to be propagated and potentially amplified throughout the video. Automating or integrating the first-frame editing process into a single, cohesive framework could address this dependency.

\paragraph{Challenges with Extreme Motion and Occlusion.} While ContextFlow is robust in many scenarios, it may face challenges with videos containing extremely rapid object or camera motion, or complex, prolonged occlusions of the edited object. In such cases, maintaining perfect temporal consistency and object integrity can be difficult for any I2V model, and our guidance mechanism may struggle to provide a stable enough signal.

\paragraph{Computational Overhead.} As a training-free method relying on a dual-path sampling process, ContextFlow is computationally intensive. As detailed in A.1, the inference time and high VRAM requirement may be a barrier for users with less powerful hardware.

\paragraph{Related work.}
We thanks these related work~\cite{sun2025gl, sun2025bs, zhang2025echomask, zhang2025semtalk, wang2025characterfactory, liu2023delving, liu2024headartist, liu2023human} and their contribution. Also, we are motivated by some related works~\cite{zhao2025unified, Zhao_2023_CVPR, Zhao_2023_ICCV_DDFM, zhu2022one, feng20254dgs, li2025gs2e, jia2019comdefend, jia2022adversarial, jia2022boosting,Z1,Z2,Z3,Z4,Z5, MAT,Z6,Z7,Z8,Z9,she2025customvideox,wang2025mint,ying2024restorerid,liu2024llm4gen,liu2024llm4gen,shen2025follow, wan2025unipaint}. Additionally, we also care about some AI-safety works~\cite{song2025idprotector,song2024anti,hui2025autoregressive,ci2024wmadapter,ci2024ringid,liu2024image,yang2024can}

\paragraph{Future Work.} Based on these limitations, we identify several promising directions for future work. First, developing a more robust framework that can handle severe motion and occlusions remains a key challenge. Second, exploring model compression or distillation techniques could significantly reduce the computational cost and improve inference speed. Finally, integrating advanced motion control mechanisms could allow for more fine-grained manipulation of the edited object’s trajectory and dynamics, moving beyond simple propagation.

{
    \small
    \bibliographystyle{ieeenat_fullname}
    \bibliography{main}
}


\end{document}